\documentclass[lettersize,journal]{IEEEtran}
\usepackage{amsmath,amsfonts}
\usepackage{algorithmic}
\usepackage{algorithm}
\usepackage{array}
\usepackage[caption=false,font=normalsize,labelfont=sf,textfont=sf]{subfig}
\usepackage{textcomp}
\usepackage{stfloats}
\usepackage{url}
\usepackage{verbatim}
\usepackage{graphicx}
\usepackage{cite}

\newcommand{\dd}{\mathop{}\!\mathrm{d}}

\usepackage{amsmath}
{
      
      \newtheorem{definition}{Definition}
\newtheorem{result}{Result}
  }
  
\usepackage{tikz}
  \usetikzlibrary{shapes.symbols}
  \usepackage{fontawesome}

\begin{document}

\title{Membership-Mappings for Practical Secure Distributed Deep Learning}
\author{Mohit~Kumar,
            Weiping~Zhang,
            Lukas~Fischer,
            and~Bernhard~Freudenthaler%
\thanks{This work was partly supported by the Austrian Research Promotion Agency (FFG) Project SMiLe (Secure Machine Learning Applications with Homomorphically Encrypted Data); FFG COMET-Modul S3AI (Security and Safety for Shared Artificial Intelligence); FFG Sub-Project PETAI (Privacy Secured Explainable and Transferable AI for Healthcare Systems); EU Horizon 2020 Project SERUMS (Securing Medical Data in Smart Patient-Centric Healthcare Systems); the BMK; the BMDW; and the Province of Upper Austria in the frame of the COMET Programme managed by FFG.}%
\thanks{M. Kumar is with the Software Competence Center Hagenberg, Austria and the Faculty of Computer Science and Electrical Engineering, University of Rostock, Germany.}
\thanks{W. Zhang is with the Northwestern Polytechnical University, Suzhouzhong Road, Suzhou, China, e-mail: zhangweiping@zjubh.com.}
\thanks{L. Fischer and B. Freudenthaler are with the Software Competence Center Hagenberg, Austria.}}

\maketitle

\begin{abstract}
This study leverages the data representation capability of fuzzy based membership-mappings for practical secure distributed deep learning using fully homomorphic encryption. The impracticality issue of secure machine (deep) learning with fully homomorphic encrypted data, arising from large computational overhead, is addressed via applying fuzzy attributes. Fuzzy attributes are induced by globally convergent and robust variational membership-mappings based local deep models. Fuzzy attributes combine the local deep models in a robust and flexible manner such that the global model can be evaluated homomorphically in an efficient manner using a boolean circuit composed of bootstrapped binary gates. The proposed method, while preserving privacy in a distributed learning scenario, remains accurate, practical, and scalable. The method is evaluated through numerous experiments including demonstrations through MNIST dataset and Freiburg Groceries Dataset. Further, a biomedical application related to mental stress detection on individuals is considered.           
\end{abstract}

\section{Introduction}
\IEEEPARstart{F}{ully} homomorphic encryption (FHE) is a solution to the privacy concerns in the cloud computing scenario. The first FHE scheme~\cite{10.1145/1536414.1536440} is based on ideal lattices and the bootstrapping procedure is introduced to reduce the noise contained in a ciphertext for allowing arbitrary computations. The bootstrapping operation is performed on a ciphertext via evaluating the decryption function homomorphically using the bootstrapping key (which is the encryption of the private decryption key under the public encryption key). Bootstrapping is the computationally most expensive part of a homomorphic encryption scheme. The theoretical breakthrough of~\cite{10.1145/1536414.1536440} was followed by several attempts to develop more practical FHE schemes. The scheme introduced in~\cite{10.1007/978-3-642-13190-5_2} uses only elementary modulo arithmetic and is homomorphic with regard to both addition and multiplication. This scheme was improved in \cite{10.1007/978-3-642-22792-9_28} with reduced public key size, extended in~\cite{10.1007/978-3-642-38348-9_20} to support encrypting and homomorphically processing a vector of plaintexts as a single ciphertext, and generalized to non-binary messages in~\cite{10.1007/978-3-662-46800-5_21}. Schemes based on a different hard problem, referred to as Learning With Errors (LWE) problem~\cite{10.1145/1060590.1060603}, were constructed and many current schemes still rely on LWE or its variants. A FHE scheme constructed in~\cite{6108154} is based solely on the standard LWE assumption that is known to be at least as hard as solving hard problems in general lattices. In a variant of the LWE problem, called ring learning with errors problem (RLWE) problem, the algebraic structure of the underlying hard problem reduces the key sizes and speeds up the homomorphic operations. A leveled fully homomorphic encryption scheme based on LWE or RLWE, without bootstrapping procedure, was proposed in~\cite{10.1145/2090236.2090262}. The ciphertexts contain a certain amount of noise for security purposes that grows with homomorphic operations. For a better management of the noise growth, \cite{10.1145/2090236.2090262} introduced a modulus switching technique where a complete ladder of moduli is used for scaling down the ciphertext to the next modulus after each multiplication. A tensoring technique for LWE-based FHE that reduced ciphertext noise growth after multiplication from quadratic to linear was introduced in~\cite{10.1007/978-3-642-32009-5_50}. As the scheme of~\cite{10.1007/978-3-642-32009-5_50} does no longer require the rescaling of the ciphertext, this scheme was called a scale-invariant fully homomorphic encryption scheme. An RLWE version of the scale-invariant scheme of~\cite{10.1007/978-3-642-32009-5_50} was created in~\cite{DBLP:journals/iacr/FanV12}. A technique for building LWE based FHE scheme called as approximate eigenvector method in which homomorphic addition and multiplication are just matrix addition and multiplication was proposed in~\cite{10.1007/978-3-642-40041-4_5}. The essence of this scheme is that the secret key is an approximate eigenvector of the ciphertext matrix and the message is the corresponding eigenvalue. Several works that followed the theoretical breakthrough of~\cite{10.1145/1536414.1536440} were aimed at improving the bootstrapping as the bootstrapping remained the bottleneck for an efficient FHE in practice. A much faster bootstrapping, based on a scheme similar to the type of~\cite{10.1007/978-3-642-40041-4_5} that allows to homomorphically compute simple bit operations and refresh (bootstrap) the resulting output in less than a second, was devised in \cite{10.1007/978-3-662-46800-5_24}. Finally, the TFHE scheme was proposed in~\cite{10.1007/978-3-662-53887-6_1,10.1007/978-3-319-70694-8_14} that features an improved bootstrapping procedure that is considerably more efficient than the previous state of the art. The TFHE scheme generalizes previous structures and schemes over the torus (i.e., the reals modulo 1) and improves the bootstrapping dramatically. For practical applications, TFHE is an open-source C/C++ library~\cite{TFHE} implementing the ring-variant of~\cite{10.1007/978-3-642-40041-4_5} together with the optimizations of~\cite{10.1007/978-3-662-46800-5_24,10.1007/978-3-662-53887-6_1,10.1007/978-3-319-70694-8_14}. TFHE library implements a very fast gate-by-gate bootstrapping and supports the homomorphic evaluation of the binary gates. The library allows to evaluate homomorphically an arbitrary boolean circuit composed of binary gates without restriction on the number of gates or on their composition, over encrypted data, without decrypting. However, the bootstrapped bit operations are still several times slower than their plaintext equivalents. 

The fuzzy systems' capability of handling uncertainties in a rigorous mathematical manner has motivated combining fuzzy theory with deep models~\cite{7377034,7482843,Zhou:2014:FDB:2583136.2583392,8054903,122268001,10.1371/journal.pone.0190831,pub.1094267909,122268001,abcd_3}. Deep neural networks outperform classical machine learning techniques in a wide range of applications but their training requires a large amount of data. The issues, such as determining the optimal model structure, requirement of large training dataset, and iterative time-consuming nature of numerical learning algorithms, are inherent to the neural networks based parametric deep models. The nonparametric approach on the other hand can be promising to address the issue of optimal choice of model structure. However, an analytical solution instead of iterative gradient-based numerical algorithms will be still desired for the learning of deep models. These motivations have led to the development of a nonparametric deep model~\cite{8888203,9216097,KUMAR20211} that is learned analytically for representing data points. The study in~\cite{8888203,9216097,KUMAR20211} introduces the concept of fuzzy-mapping which is about representing mappings through a fuzzy set such that the dimension of membership function increases with an increasing data size. A relevant result is that a deep autoencoder model formed via a composition of finite number of nonparametric fuzzy-mappings can be learned analytically via variational optimization technique. However, \cite{8888203,9216097,KUMAR20211} didn't provide a formal mathematical framework for the conceptualization of so-called fuzzy-mapping. The study in~\cite{10.1007/978-3-030-87101-7_13} provides to fuzzy-mapping a measure-theoretic conceptualization and refers it to as \emph{membership-mapping}. Further, the membership-mapping could serve as the building block of deep models~\cite{10.1007/978-3-030-87101-7_14}. An alternative idea of deep autoencoder, that consists of layers such that each layer learns data representation at certain abstraction level through a membership-mappings based autoencoder, is introduced in~\cite{10.1007/978-3-030-87101-7_14} for data representation learning.

The aim of this study is to develop a methodology for practical secure distributed deep learning using fully homomorphic encryption. The machine (deep) learning with fully homomorphically encrypted data remains impractical due to the large computational overhead. Thus, we address in this study the impracticality issue of secure distributed deep learning via
\begin{enumerate}
\item leveraging data representation learning capability of globally convergent and robust membership-mappings to build local deep models,
\item using local deep models to induce fuzzy attributes such that defined fuzzy attributes learn data representation,
\item combining local deep models in a robust and flexible manner by means of fuzzy attributes and fuzzy rules to define a global model, 
\item defining the global model in such a way that the global model can be evaluated homomorphically in an efficient manner using a boolean circuit composed of bootstrapped binary gates,
\item implementing very fast gate-by-gate bootstrapping to homomorphically evaluate the global model (that combines the distributed local models) to predict the output. 
\end{enumerate}

The proposed approach to secure distributed deep learning is novel. To the best knowledge of the authors, this is the first study to apply fuzzy attributes, which are induced by globally convergent and robust variational membership-mappings based local deep models, for an efficient homomorphic evaluation of the global model. The idea of using fuzzy sets and fuzzy rules to aggregate the local private deep models for building the global model was also considered in~\cite{KUMAR202187}, however, under differential privacy framework. Differential privacy preserves the privacy of the training dataset via adding random noise to ensure that an adversary can not infer any single data instance by observing model parameters or model outputs. The amount of noise depends upon the value of privacy-loss bound. A major limitation of the differential privacy is that a sufficiently low value of privacy-loss bound results in a loss of accuracy. Moreover, it is not clear how to practically choose the value of privacy-loss bound. FHE approach on the other hand does not lead to the loss of accuracy, however, requires a large computational time. 

The data representation learning capability of membership-mappings is central to our methodology. Although~\cite{10.1007/978-3-030-87101-7_13} provided an algorithm for the variational learning of membership-mappings via following the approach of~\cite{8888203,9216097,KUMAR20211}, there remains the following two limitations: 
\begin{enumerate}
\item there is no mathematical proof regarding the convergence of the learning algorithm, and 
\item there is no mathematical analysis on the robustness of the learning algorithm.  
\end{enumerate}
This study addresses these two limitations and presents a more simple and elegant estimation algorithm for the variational learning of membership-mappings. A convergence analysis is carried out via deriving a sufficient condition for the convergence of the estimation algorithm. The convergence analysis allows to provide a globally convergent algorithm for the variational learning of membership-mappings based deep models. Further, it is shown that the learning algorithm provides a robust estimation of model parameters via solving a min-max estimation problem. The proposed method for secure distributed deep learning is implemented using MATLAB R2017b and TFHE C/C++ library~\cite{TFHE}. Experiments have been performed to evaluate the method (in-terms of accuracy and computational time) on MNIST dataset, Freiburg Groceries Dataset, and a biomedical dataset consisting of heart rate interval measurements of different subjects. Further, the scalability of the method as the number of parties participating in collaborative learning increases is investigated.    

The paper is organized into sections. Section~\ref{sec_review} reviews the membership-mappings from previous works. An estimation algorithm for the variational learning of membership-mappings is provided in Section~\ref{sec_variational_learning}. The convergence and robustness issues have been addressed in Section~\ref{sec_convergence_robustness}. Section~\ref{sec_application} considers the application of membership-mappings to the secure distributed deep learning problem. The experimental validation of the method is provided in Section~\ref{sec_experiments}. Finally, the concluding remarks are stated in Section~\ref{sec_conclusion}.

\section{Review of Membership-Mappings}\label{sec_review}
This section is dedicated to the review of variational membership-mappings from previous works~\cite{10.1007/978-3-030-87101-7_13,10.1007/978-3-030-87101-7_14}.
\subsection{Notations and Definitions}
\begin{itemize}
\item Let $n,N,p,M \in \mathbb{N}$. 
\item Let $\mathcal{B}(\mathbb{R}^N)$ denote the \emph{Borel $\sigma-$algebra} on $\mathbb{R}^N$, and let $\lambda^N$ denote the \emph{Lebesgue measure} on $\mathcal{B}(\mathbb{R}^N)$. 
\item Let $(\mathcal{X}, \mathcal{A} , \rho)$ be a probability space with unknown probability measure $\rho$. 
\item Let us denote by $\mathcal{S}$ the set of finite samples of data points drawn i.i.d. from $\rho$, i.e.,
\begin{IEEEeqnarray}{rCl}  
\mathcal{S} & := &  \{ (x^i  \sim \rho )_{i=1}^N \; | \; N \in \mathbb{N} \}.
\end{IEEEeqnarray} 
\item For a sequence $\mathrm{x} = (x^1,\cdots,x^N) \in \mathcal{S}$, let $|\mathrm{x}|$ denote the cardinality i.e. $|\mathrm{x}| = N$.     
\item If $\mathrm{x} = (x^1, \cdots, x^N),\; \mathrm{a}= (a^1, \cdots, a^M) \in \mathcal{S}$, then $\mathrm{x} \wedge \mathrm{a}$ denotes the concatenation of the sequences $\mathrm{x}$ and $\mathrm{a}$, i.e., $\mathrm{x} \wedge \mathrm{a} = (x^1, \ldots, x^N, a^1, \ldots, a^M)$.
\item Let us denote by $\mathbb{F}(\mathcal{X})$ the set of $\mathcal{A}$-$\mathcal{B}(\mathbb{R})$ measurable functions $f:\mathcal{X} \rightarrow \mathbb{R}$, i.e.,
\begin{IEEEeqnarray}{rCl}  
\mathbb{F}(\mathcal{X}) & := &  \{ f:\mathcal{X} \rightarrow \mathbb{R}  \; | \;  \mbox{$f$ is $\mathcal{A}$-$\mathcal{B}(\mathbb{R})$ measurable}\}.
\end{IEEEeqnarray} 
\item For convenience, the values of a function $f \in \mathbb{F}(\mathcal{X})$ at points in the collection $\mathrm{x} = (x^1,\cdots,x^N)$ are represented as $f(\mathrm{x})=(f(x^1),\cdots,f(x^N))$.
\item Given two $\mathcal{B}(\mathbb{R}^N)-\mathcal{B}(\mathbb{R})$ measurable mappings, $g:\mathbb{R}^N \rightarrow \mathbb{R}$ and $\mu:\mathbb{R}^N \rightarrow \mathbb{R}$, the weighted average of $g(\mathrm{y})$ over all $\mathrm{y} \in \mathbb{R}^{N}$, with $\mu(\mathrm{y})$ as the weighting function, is computed as   
\begin{IEEEeqnarray}{rCl}
\label{eq_738118.427179} \left< g \right>_{\mu}& := & \frac{1}{ \int_{\mathbb{R}^{N}}  \mu(\mathrm{y})\, \dd\lambda^{N}(\mathrm{y})} \int_{\mathbb{R}^{N}} g(\mathrm{y}) \mu(\mathrm{y})\, \dd \lambda^{N}(\mathrm{y}). \IEEEeqnarraynumspace
\end{IEEEeqnarray} 
\item For a sequence $\mathrm{x} \in \mathcal{S}$, assume that a membership function $\zeta_{\mathrm{x}}:\mathbb{R}^{|\mathrm{x}|} \rightarrow [0,1]$ satisfies the following properties:
\begin{itemize}
    \item $\zeta_{\mathrm{x}}(\mathrm{y}) > 0$ for all $\mathrm{y} \in \mathbb{R}^{|\mathrm{x}|}$, i.e.,
\begin{IEEEeqnarray}{rCl}
    \label{eq:supp}
     \mbox{supp}[\zeta_{\mathrm{x}}] & = & \mathbb{R}^{|\mathrm{x}|}.
\end{IEEEeqnarray}   
    \item the functions $\zeta_{\mathrm{x}}$ are absolutely continuous and Lebesgue integrable over the whole domain such that for all $\mathrm{x}\in \mathcal{S}$ we have
     \begin{eqnarray}
    \label{eq:positive}
   0 < \int_{\mathbb{R}^{|\mathrm{x}|}} \zeta_{\mathrm{x}}\, \dd \lambda^{|\mathrm{x}|}  < \infty.
   \end{eqnarray}
   \item the membership function induced probability measures $\mathbb{P}_{\zeta_{\mathrm{x}}}$, defined on any $A \in \mathcal{B}(\mathbb{R}^{|\mathrm{x}|})$, as
\begin{IEEEeqnarray}{rCl}
\mathbb{P}_{\zeta_{\mathrm{x}}}(A) & := &  \frac{1}{ \int_{\mathbb{R}^{|\mathrm{x}|}} \zeta_{\mathrm{x}}\, \dd \lambda^{|\mathrm{x}|}} \int_{A} \zeta_{\mathrm{x}}\, \dd\lambda^{|\mathrm{x}|}
\end{IEEEeqnarray}  
are consistent in the sense that for all $\mathrm{x},\;\mathrm{a} \in \mathcal{S}$:
\begin{IEEEeqnarray}{rCl}
\label{eq_738083.390026} \mathbb{P}_{\zeta_{\mathrm{x} \wedge \mathrm{a}}}(A \times \mathbb{R}^{|\mathrm{a}|}) & = & \mathbb{P}_{\zeta_{\mathrm{x}}}(A). 
\end{IEEEeqnarray}  
\end{itemize}
The collection of membership functions satisfying aforementioned assumptions is denoted by 
\begin{IEEEeqnarray}{rCl}
\Theta & := & \{ \zeta_{\mathrm{x}}:\mathbb{R}^{|\mathrm{x}|} \rightarrow [0,1] \; | \; (\ref{eq:supp}),  (\ref{eq:positive}), (\ref{eq_738083.390026}),\; \mathrm{x} \in \mathcal{S}\}. \IEEEeqnarraynumspace
\end{IEEEeqnarray}
\end{itemize}   
\begin{definition}[Student-t Membership-Mapping~\cite{10.1007/978-3-030-87101-7_13}]\label{def_student_t_set_membership_mapping}
A Student-t membership-mapping, $\mathcal{F} \in \mathbb{F}(\mathcal{X})$, is a mapping with input space $\mathcal{X} = \mathbb{R}^n$ and a membership function $\zeta_{\mathrm{x}} \in \Theta$ that is Student-t like:
\begin{IEEEeqnarray}{rCl}
\label{eq_student_t_membership} 
\label{eq_738098.751419}\zeta_{\mathrm{x}}(\mathrm{y}) & = & \left(1 + \frac{1}{\nu - 2}  \left( \mathrm{y} - \mathrm{m}_{\mathrm{y}} \right)^T K^{-1}_{\mathrm{x}\mathrm{x}} \left( \mathrm{y}- \mathrm{m}_{\mathrm{y}}\right) \right)^{-\frac{\nu+|\mathrm{x}|}{2}} \IEEEeqnarraynumspace
\end{IEEEeqnarray} 
where $\mathrm{x} \in \mathcal{S}$, $\mathrm{y} \in \mathbb{R}^{|\mathrm{x}|}$, $\nu \in \mathbb{R}_{+}\setminus [0,2]$ is the degrees of freedom, $\mathrm{m}_{\mathrm{y}} \in \mathbb{R}^{|\mathrm{x}|}$ is the mean vector, and $K_{\mathrm{x}\mathrm{x}} \in \mathbb{R}^{|\mathrm{x}| \times |\mathrm{x}|}$ is the covariance matrix with its $(i,j)-$th element given as 
\begin{IEEEeqnarray}{rCl}
\label{738026.844153}  (K_{\mathrm{x}\mathrm{x}})_{i,j} & = & kr(x^i,x^j) 
 \end{IEEEeqnarray}  
where $kr: \mathbb{R}^n \times \mathbb{R}^n \rightarrow \mathbb{R}$ is a positive definite kernel function defined as 
\begin{IEEEeqnarray}{rCl}
\label{eq_membership1003_3} kr(x^{i},x^{j}) & = &  \sigma^2 \exp \left(-0.5\sum_{k = 1}^{n} w_{k} \left |  x^{i}_k - x^{j}_k \right |^2\right)
 \end{IEEEeqnarray}  
where $x_k^i$ is the $k-$th element of $x^i$, $\sigma^2$ is the variance parameter, and $w_{k} \geq 0$ (for $k \in \{1,\cdots,n\}$).
\end{definition}
\subsection{Conditionally Deep Autoencoders}
This subsection reviews the conditionally deep models~\cite{10.1007/978-3-030-87101-7_14} formed by the compositions of membership-mappings.    
\begin{definition}[Membership-Mapping Autoencoder~\cite{10.1007/978-3-030-87101-7_14}]\label{def_SFMA}
A membership-mapping autoencoder, $\mathcal{G}:\mathbb{R}^p \rightarrow \mathbb{R}^p$, maps an input vector $y \in \mathbb{R}^p$ to $\mathcal{G}(y) \in \mathbb{R}^p$ such that 
 \begin{IEEEeqnarray}{rCl}
\label{eq_added_before_publication_1}  \mathcal{G}(y) &    \overset{\underset{\mathrm{def}}{}}{=} &  \left[\begin{IEEEeqnarraybox*}[][c]{,c/c/c,}  \mathcal{F}_1(Py) & \cdots &  \mathcal{F}_p(Py)
 \end{IEEEeqnarraybox*} \right]^T, 
\end{IEEEeqnarray} 
where $\mathcal{F}_j$ ($j \in \{1,2,\cdots,p\}$) is a Student-t membership-mapping, $P \in \mathbb{R}^{n \times p} (n \leq p)$ is a matrix such that the product $Py$ is a lower-dimensional encoding for $y$. 
\end{definition}
\begin{definition}[Conditionally Deep Membership-Mapping Autoencoder (CDMMA)~\cite{10.1007/978-3-030-87101-7_14}]\label{def_deep_autoencoder}
A conditionally deep membership-mapping autoencoder, $\mathcal{D}:\mathbb{R}^p \rightarrow \mathbb{R}^p$, maps a vector $y \in \mathbb{R}^p$ to $\mathcal{D}(y) \in \mathbb{R}^p$ through a nested composition of finite number of membership-mapping autoencoders such that
 \begin{IEEEeqnarray}{rCl}
 y^l & = & (\mathcal{G}_l \circ \cdots \circ \mathcal{G}_2\circ \mathcal{G}_1)(y), \; \forall l \in \{1,2,\cdots,L \}\\
 l^* & = & \arg\;\min_{l \: {\in} \: \{1,2,\cdots,L \}}\; \| y -  y^l \|^2 \\
  \mathcal{D}(y) & = & y^{l^*},
\end{IEEEeqnarray}  
where $\mathcal{G}_l(\cdot)$ is a membership-mapping autoencoder (Definition~\ref{def_SFMA}).
\end{definition}
\begin{definition}[A Wide CDMMA~\cite{10.1007/978-3-030-87101-7_14}]\label{def_wide_deep_autoencoder}
A wide CDMMA, $\mathcal{WD}:\mathbb{R}^p \rightarrow \mathbb{R}^p$, maps a vector $y \in \mathbb{R}^p$ to $\mathcal{WD}(y) \in \mathbb{R}^p$ through a parallel composition of $S$  ($S \in \mathcal{Z}_+$) number of CDMMAs such that
 \begin{IEEEeqnarray}{rCl}
\label{eq_738125.489500} \mathcal{WD}(y) & = & \mathcal{D}_{s^*}(y)\\
 s^* & = & \arg\;\min_{s \in \{1,2,\cdots,S \}}\; \| y -  \mathcal{D}_s(y)  \|^2,
 \end{IEEEeqnarray}  
where $\mathcal{D}_s(y)$ is the output of $s-$th CDMMA. 
\end{definition}

\section{Variational Learning of Membership-Mappings}\label{sec_variational_learning}
\subsection{A Modeling Scenario}
Given a dataset $\{(x^i,y^i)\;|\; x^i \in \mathbb{R}^n,\;y^i \in \mathbb{R}^p,\; i \in \{1,\cdots,N \} \}$, it is assumed that there exist zero-mean Student-t membership-mappings $\mathcal{F}_1, \cdots, \mathcal{F}_p \in \mathbb{F}(\mathbb{R}^n)$ such that
\begin{IEEEeqnarray}{rCl}
\label{eq_738118.641846} y^i &\approx & \left[\begin{IEEEeqnarraybox*}[][c]{,c/c/c,} \mathcal{F}_1(x^i)  & \cdots & \mathcal{F}_p(x^i) \end{IEEEeqnarraybox*} \right]^T.
\end{IEEEeqnarray} 
For $ j \in \{1,2,\cdots,p\}$, define
\begin{IEEEeqnarray}{rCl}
\label{eq_y_j_vec2000} \mathrm{y}_j & = & \left[\begin{IEEEeqnarraybox*}[][c]{,c/c/c,}y_j^1 & \cdots &y_j^N\end{IEEEeqnarraybox*} \right]^T \in \mathbb{R}^N\\
\label{eq_f_j_vec} \mathrm{f}_j & = & \left[\begin{IEEEeqnarraybox*}[][c]{,c/c/c,}\mathcal{F}_j(x^1) & \cdots & \mathcal{F}_j(x^N)\end{IEEEeqnarraybox*} \right]^T \in \mathbb{R}^N 
 \end{IEEEeqnarray}  
where $y_j^i$ denotes the $j-$th element of $y^i$. A set of auxiliary inducing points, $\mathrm{a}  =  \{a^{m} \in \mathbb{R}^{n}\; | \; m \in\{1,\cdots,M\} \}$, is introduced to define 
\begin{IEEEeqnarray}{rCl}
\label{738018.270413}\mathrm{u}_j =  \left[\begin{IEEEeqnarraybox*}[][c]{,c/c/c,}\mathcal{F}_j(a^{1}) & \cdots & \mathcal{F}_j(a^{M})\end{IEEEeqnarraybox*} \right]^T \in \mathbb{R}^M.
\end{IEEEeqnarray}
\subsection{Membership Functional Representation Approach}
\begin{definition}[Interpolation Based Representation]\label{def_738117.592248}
It follows from~\cite{10.1007/978-3-030-87101-7_13} that $\mathrm{f}_j$, based upon an interpolation on the auxiliary-outputs $\mathrm{u}_j $, is represented by means of a membership function, $\mu_{\mathrm{f}_j;\mathrm{u}_j}:\mathbb{R}^N \rightarrow [0,1]$, as
\begin{IEEEeqnarray}{rCl}
\label{eq_pf_1001_student_t} \lefteqn{\left( \mu_{\mathrm{f}_j;\mathrm{u}_j}(\tilde{\mathrm{f}}_j) \right)^{-\frac{2}{\nu+M+N}}  =  1 + } \\
\nonumber &&  \frac{(\tilde{\mathrm{f}}_j -  \bar{m}_{\mathrm{f}_j})^T  \left( \frac{\nu + (\mathrm{u}_j)^T K_{\mathrm{a}\mathrm{a}}^{-1} \mathrm{u}_j  - 2}{\nu + M - 2} \bar{K}_{\mathrm{x}\mathrm{x}} \right)^{-1}(\tilde{\mathrm{f}}_j -  \bar{m}_{\mathrm{f}_j})}{\nu + M - 2}  \\
\label{eq_pf_1002}  \bar{m}_{\mathrm{f}_j} & = &  K_{\mathrm{x}\mathrm{a}} K_{\mathrm{a}\mathrm{a}}^{-1}  \mathrm{u}_j   \\
\label{eq_pf_1003} \bar{K}_{\mathrm{x}\mathrm{x}} & = & K_{\mathrm{x}\mathrm{x}} - K_{\mathrm{x}\mathrm{a}} K_{\mathrm{a}\mathrm{a}}^{-1} K_{\mathrm{x}\mathrm{a}}^T,
\end{IEEEeqnarray}   
where $K_{\mathrm{a}\mathrm{a}} \in \mathbb{R}^{M \times M}$ and $K_{\mathrm{x}\mathrm{a}}  \in \mathbb{R}^{N \times M}$ are matrices with their $(i,j)-$th elements given as
\begin{IEEEeqnarray}{rCl}
\label{eq_membership1004_2} \left( K_{\mathrm{a}\mathrm{a}} \right)_{i,j} & = & kr(a^{i},a^{j}) \\
\label{eq_738497.4922} \left( K_{\mathrm{x}\mathrm{a}} \right)_{i,j} & = & kr(x^{i},a^{j})
 \end{IEEEeqnarray} 
where $kr: \mathbb{R}^n \times \mathbb{R}^n \rightarrow \mathbb{R}$ is a positive definite kernel function defined as in (\ref{eq_membership1003_3}).    
\end{definition}
\begin{definition}[Representation of Data $\mathrm{y}_j$ for Given Mappings Output $\mathrm{f}_j$]\label{def_738494.462}
$\mathrm{y}_j$, for given $\mathrm{f}_j$, is represented by means of a membership function, $\mu_{\mathrm{y}_j;\mathrm{f}_j}:\mathbb{R}^N \rightarrow [0,1]$, as
\begin{IEEEeqnarray}{rCl}
\label{eq_membership1002} \mu_{\mathrm{y}_j;\mathrm{f}_j}(\tilde{\mathrm{y}}_j) & = & \exp\left(- 0.5 \beta \|  \tilde{\mathrm{y}}_j -  \mathrm{f}_j \|^2 \right)
\end{IEEEeqnarray}    
where $\beta > 0$ is the precision value. 
\end{definition}
\begin{definition}[Representation of Data $\mathrm{y}_j$ for Fixed Auxiliary-Outputs $\mathrm{u}_j$]
$\mathrm{y}_j$, for given $\mathrm{u}_j$, is represented by means of a membership function, $\mu_{\mathrm{y}_j;\mathrm{u}_j}:\mathbb{R}^N \rightarrow [0,1]$, as
\begin{IEEEeqnarray}{rCl}
\label{eq_738494.4689}   \mu_{\mathrm{y}_j;\mathrm{u}_j}(\tilde{\mathrm{y}}_j)  & \propto  & \exp \left( \left< \log ( \mu_{\mathrm{y}_j;\mathrm{f}_j}(\tilde{\mathrm{y}}_j) )  \right >_{\mu_{\mathrm{f}_j;\mathrm{u}_j}} \right) \IEEEeqnarraynumspace
\end{IEEEeqnarray} 
where $\mu_{\mathrm{y}_j;\mathrm{f}_j}$ is given by (\ref{eq_membership1002}), $\mu_{\mathrm{f}_j;\mathrm{u}_j}$ is defined as in (\ref{eq_pf_1001_student_t}), and $<\cdot>_{\cdot}$ is the averaging operation as defined in (\ref{eq_738118.427179}). Thus, $\mu_{\mathrm{y}_j;\mathrm{u}_j}$ is obtained from $\log( \mu_{\mathrm{y}_j;\mathrm{f}_j})$ after averaging out the variables  $\mathrm{f}_j$ using its membership function. It can be shown that
\begin{IEEEeqnarray}{rCl}
\label{eq_log_membership2} \lefteqn{  \mu_{\mathrm{y}_j;\mathrm{u}_j}(\tilde{\mathrm{y}}_j)  \propto  \exp \left(  - 0.5\beta \| \tilde{\mathrm{y}}_j\|^2  + (\mathrm{u}_j)^T \hat{K}_{\mathrm{u}_j}^{-1} \hat{m}_{\mathrm{u}_j }(\tilde{\mathrm{y}}_j) \right.} \\
\nonumber && \left. - 0.5 (\mathrm{u}_j)^T \hat{K}_{\mathrm{u}_j}^{-1} \mathrm{u}_j + 0.5 (\mathrm{u}_j)^T  K_{\mathrm{a}\mathrm{a}}^{-1}  \mathrm{u}_j +  \{/ (\tilde{\mathrm{y}}_j,\mathrm{u}_j)\} \right) \IEEEeqnarraynumspace
\end{IEEEeqnarray}   
where $\hat{K}_{\mathrm{u}_j}$, $\hat{m}_{\mathrm{u}_j}(\tilde{\mathrm{y}}_j)$ are given as
\begin{IEEEeqnarray}{rCl}
\label{eq_hat_K_u_1000}(\hat{K}_{\mathrm{u}_j})^{-1} & = & K_{\mathrm{a}\mathrm{a}}^{-1}  + \beta K_{\mathrm{a}\mathrm{a}}^{-1}  K_{\mathrm{x}\mathrm{a}}^T K_{\mathrm{x}\mathrm{a}}  K_{\mathrm{a}\mathrm{a}}^{-1}  \\
\nonumber &&  {+}\: \beta \frac{\text{tr}(K_{\mathrm{x}\mathrm{x}} - K_{\mathrm{a}\mathrm{a}}^{-1}  K_{\mathrm{x}\mathrm{a}}^T K_{\mathrm{x}\mathrm{a}} )}{\nu + M - 2} K_{\mathrm{a}\mathrm{a}}^{-1}  \\
\label{eq_hat_m_u_1000} \hat{m}_{\mathrm{u}_j}(\tilde{\mathrm{y}}_j) & = & \beta \hat{K}_{\mathrm{u}_j} K_{\mathrm{a}\mathrm{a}}^{-1} K_{\mathrm{x}\mathrm{a}}^T \tilde{\mathrm{y}}_j,  
\end{IEEEeqnarray}  
and $ \{/ (\tilde{\mathrm{y}}_j,\mathrm{u}_j)\}$ represents all those terms which are independent of both $\tilde{\mathrm{y}}_j$ and $\mathrm{u}_j$. The constant of proportionality in (\ref{eq_log_membership2}) is chosen to exclude $ (\tilde{\mathrm{y}}_j,\mathrm{u}_j)-$independent terms in the expression for $\mu_{\mathrm{y}_j;\mathrm{u}_j}$, i.e., 
\begin{IEEEeqnarray}{rCl}
\label{eq_scch_1} \lefteqn{ \mu_{\mathrm{y}_j;\mathrm{u}_j}(\tilde{\mathrm{y}}_j)   =  \exp\left(  (\mathrm{u}_j)^T \hat{K}_{\mathrm{u}_j}^{-1} \hat{m}_{\mathrm{u}_j }(\tilde{\mathrm{y}}_j) \right.}  \IEEEeqnarraynumspace \\
\nonumber && \left.  - 0.5 (\mathrm{u}_j)^T \hat{K}_{\mathrm{u}_j}^{-1} \mathrm{u}_j + 0.5 (\mathrm{u}_j)^T  K_{\mathrm{a}\mathrm{a}}^{-1}  \mathrm{u}_j - 0.5\beta \| \tilde{\mathrm{y}}_j\|^2 \right). 
\end{IEEEeqnarray}          
\end{definition}
\begin{definition}[Data-Model]\label{def_avg_avg_membership}
$\mathrm{y}_j$ is represented by means of a membership function, $\mu_{\mathrm{y}_j} : \mathbb{R}^N \rightarrow [0,1]$, as 
\begin{IEEEeqnarray}{rCl}
\label{eq_scch_2}  \mu_{\mathrm{y}_j}(\tilde{\mathrm{y}}_j) & \propto & \exp \left( \left<  \log( \mu_{\mathrm{y}_j;\mathrm{u}_j} (\tilde{\mathrm{y}}_j) )\right>_{\mu_{\mathrm{u}_j}} \right)
\end{IEEEeqnarray} 
where $ \mu_{\mathrm{y}_j;\mathrm{u}_j}$ is given by (\ref{eq_scch_1}) and $\mu_{\mathrm{u}_j}:  \mathbb{R}^M \rightarrow [0,1]$ is a membership function representing $\mathrm{u}_j$. Thus, $\mu_{\mathrm{y}_j}$ is obtained from $\log(\mu_{\mathrm{y}_j;\mathrm{u}_j})$ after averaging out the auxiliary-outputs $\mathrm{u}_j$ using membership function $\mu_{\mathrm{u}_j}$. 
\end{definition}   
\subsection{Variational Optimization of Data-Model}
The data model (\ref{eq_scch_2}) involves the membership function $\mu_{\mathrm{u}_j}$. To determine $\mu_{\mathrm{u}_j}$ for a given $\mathrm{y}_j$, $\log(\mu_{\mathrm{y}_j}(\mathrm{y}_j))$ is maximized w.r.t. $\mu_{\mathrm{u}_j}$ around an initial guess. The zero-mean Gaussian membership function with covariance as equal to $K_{\mathrm{a}\mathrm{a}}$ is taken as the initial guess. It follows from (\ref{eq_scch_2}) that maximization of $\log(\mu_{\mathrm{y}_j}(\mathrm{y}_j))$ is equivalent to the maximization of $\left<  \log( \mu_{\mathrm{y}_j;\mathrm{u}_j}(\mathrm{y}_j) )\right>_{\mu_{\mathrm{u}_j}}$.  
\begin{result}\label{result_optimal_u}
The solution of following maximization problem:
\begin{IEEEeqnarray}{rCl}
\mu^*_{\mathrm{u}_j}   & = &  \arg \: \max_{\displaystyle \mu_{\mathrm{u}_j}} \: \left[ \left<  \log( \mu_{\mathrm{y}_j;\mathrm{u}_j}(\mathrm{y}_j) )\right>_{\mu_{\mathrm{u}_j}} \right. \\
\nonumber && \left. - \left< \log(\frac{ \mu_{\mathrm{u}_j}(\mathrm{u}_j)}{\exp\left(-0.5 (\mathrm{u}_j)^T K_{\mathrm{a}\mathrm{a}}^{-1} \mathrm{u}_j  \right)} ) \right>_{ \mu_{\mathrm{u}_j}} \right] 
\end{IEEEeqnarray} 
under the fixed integral constraint: 
\begin{IEEEeqnarray}{rCl}
\int_{\mathbb{R}^M} \mu_{\mathrm{u}_j}  \, \dd \lambda^M   =  C_{\mathrm{u}_j} > 0
\end{IEEEeqnarray} 
 where the value of $C_{\mathrm{u}_j}$ is so chosen such that the maximum possible values of $\mu^*_{\mathrm{u}_j} $ remain as equal to unity, is given as
\begin{IEEEeqnarray}{rCl}
\label{eq_q_u_vec_optimal} \lefteqn{\mu^*_{\mathrm{u}_j}(\mathrm{u}_j)   = } \\
\nonumber && \exp\left(- 0.5 \left(\mathrm{u}_j - \hat{m}_{\mathrm{u}_j}(\mathrm{y}_j)\right)^T \hat{K}_{\mathrm{u}_j}^{-1}\left(\mathrm{u}_j - \hat{m}_{\mathrm{u}_j}(\mathrm{y}_j)\right) \right)  
\end{IEEEeqnarray}
where $\hat{K}_{\mathrm{u}_j}$ and $\hat{m}_{\mathrm{u}_j}$ are given by (\ref{eq_hat_K_u_1000}) and (\ref{eq_hat_m_u_1000}) respectively. The solution of the optimization problem results in  
\begin{IEEEeqnarray}{rCl}
 \label{eq_satguru_8} \lefteqn{\mu_{\mathrm{y}_j}(\tilde{\mathrm{y}}_j)   \propto \exp \left(\{/ (\mathrm{y}_j,\tilde{\mathrm{y}}_j)\}   \right.} \\
\nonumber && \left. {-}\: 0.5\beta\left\{\| \tilde{\mathrm{y}}_j\|^2  - 2 \left(\hat{m}_{\mathrm{u}_j }(\mathrm{y}_j)\right)^T K_{\mathrm{a}\mathrm{a}}^{-1} K_{\mathrm{x}\mathrm{a}}^T \tilde{\mathrm{y}}_j \right. \right. \\
\nonumber && \left. \left.  {+}\:  \left(\hat{m}_{\mathrm{u}_j }(\mathrm{y}_j)\right)^TK_{\mathrm{a}\mathrm{a}}^{-1}  K_{\mathrm{x}\mathrm{a}}^T K_{\mathrm{x}\mathrm{a}} K_{\mathrm{a}\mathrm{a}}^{-1}\hat{m}_{\mathrm{u}_j }(\mathrm{y}_j)  \right. \right. \\
\nonumber  && \left.  \left.  {+}\: \frac{\text{tr}(K_{\mathrm{x}\mathrm{x}} - K_{\mathrm{a}\mathrm{a}}^{-1}  K_{\mathrm{x}\mathrm{a}}^T K_{\mathrm{x}\mathrm{a}} )}{\nu + M - 2}  \left(\hat{m}_{\mathrm{u}_j }(\mathrm{y}_j)\right)^T K_{\mathrm{a}\mathrm{a}}^{-1}\hat{m}_{\mathrm{u}_j }(\mathrm{y}_j)  \right \} \right) 
\end{IEEEeqnarray} 
where $\{/ (\mathrm{y}_j,\tilde{\mathrm{y}}_j)\}$ represents all $(\mathrm{y}_j,\tilde{\mathrm{y}}_j)-$independent terms.
\end{result}
\begin{IEEEproof}
The proof is similar to as that of Result~2 in \cite{8888203}.
\end{IEEEproof}
The constant of proportionality in (\ref{eq_satguru_8}) is chosen to exclude $(\mathrm{y}_j,\tilde{\mathrm{y}}_j)-$independent terms resulting in
\begin{IEEEeqnarray}{rCl}
 \label{eq_738499.7546} \lefteqn{\mu_{\mathrm{y}_j}(\tilde{\mathrm{y}}_j)  =   \exp \left( - \frac{\beta}{2} \left\{\| \tilde{\mathrm{y}}_j\|^2 - 2 \left(\hat{m}_{\mathrm{u}_j }(\mathrm{y}_j)\right)^T K_{\mathrm{a}\mathrm{a}}^{-1} K_{\mathrm{x}\mathrm{a}}^T \tilde{\mathrm{y}}_j \right. \right.} \IEEEeqnarraynumspace \\
 \nonumber && \left. \left.  +  \left(\hat{m}_{\mathrm{u}_j }(\mathrm{y}_j)\right)^TK_{\mathrm{a}\mathrm{a}}^{-1}  K_{\mathrm{x}\mathrm{a}}^T K_{\mathrm{x}\mathrm{a}} K_{\mathrm{a}\mathrm{a}}^{-1}\hat{m}_{\mathrm{u}_j }(\mathrm{y}_j)  \right. \right. \\
\nonumber  && \left.  \left.  + \left(\hat{m}_{\mathrm{u}_j }(\mathrm{y}_j)\right)^T\frac{\text{tr}(K_{\mathrm{x}\mathrm{x}} - K_{\mathrm{a}\mathrm{a}}^{-1}  K_{\mathrm{x}\mathrm{a}}^T K_{\mathrm{x}\mathrm{a}} )}{\nu + M - 2} K_{\mathrm{a}\mathrm{a}}^{-1}\hat{m}_{\mathrm{u}_j }(\mathrm{y}_j)  \right \} \right).
\end{IEEEeqnarray}
\subsection{Estimation of Membership-Mapping Parameters}
 \begin{definition}[Averaged Estimation of Membership-Mapping Output]\label{def_satguru_1}
$\mathcal{F}_j(x^{i})$ (which is the $i-$th element of vector $\mathrm{f}_j$~(\ref{eq_f_j_vec})) can be estimated as
 \begin{IEEEeqnarray}{rCl} 
\widehat{ \mathcal{F}_j(x^{i})} & := &  \left<  \left< (\mathrm{f}_j)_i \right>_{\mu_{\mathrm{f}_j;\mathrm{u}_j } }  \right>_{ \mu^*_{\mathrm{u}_j}}
\end{IEEEeqnarray} 
where $(\mathrm{f}_j)_i$ denotes the $i-$th element of $\mathrm{f}_j$, $\mu_{\mathrm{f}_j;\mathrm{u}_j } $ is defined as in~(\ref{eq_pf_1001_student_t}), and $\mu^*_{\mathrm{u}_j}$ is the optimal membership function (\ref{eq_q_u_vec_optimal}) representing $\mathrm{u}_j$. That is, $\mathcal{F}_j(x^{i})$, being a function of $\mathrm{u}_j$, is averaged over $\mathrm{u}_j$ for an estimation.
\end{definition}

Let $G(x) \in \mathbb{R}^{1 \times M}$ be a vector-valued function defined as
\begin{IEEEeqnarray}{rCl}
\label{eq_738495.5497} G(x)& := &  \left[\begin{IEEEeqnarraybox*}[][c]{,c/c/c,}kr(x,a^{1}) & \cdots & kr(x,a^{M}) \end{IEEEeqnarraybox*} \right]
 \end{IEEEeqnarray} 
where $kr: \mathbb{R}^n \times \mathbb{R}^n \rightarrow \mathbb{R}$ is defined as in (\ref{eq_membership1003_3}). It is shown in Appendix \ref{appendix_738242.626259} that
\begin{IEEEeqnarray}{rCl}
\label{eq_satguru_12} \widehat{ \mathcal{F}_j(x^{i})} &  = & G(x^i) \left( K_{\mathrm{x}\mathrm{a}}^T K_{\mathrm{x}\mathrm{a}}   +  \tau K_{\mathrm{a}\mathrm{a}} + \beta^{-1} K_{\mathrm{a}\mathrm{a}}\right)^{-1}  K_{\mathrm{x}\mathrm{a}}^T \mathrm{y}_j \IEEEeqnarraynumspace
 \end{IEEEeqnarray} 
where $\tau$ is given as 
\begin{IEEEeqnarray}{rCl}
 \label{eq_738566.838771} \tau & := & \frac{\text{tr}(K_{\mathrm{x}\mathrm{x}} - K_{\mathrm{a}\mathrm{a}}^{-1}  K_{\mathrm{x}\mathrm{a}}^T K_{\mathrm{x}\mathrm{a}} )}{\nu+ M - 2}.
 \end{IEEEeqnarray}
Define a vector $\alpha_j \in \mathbb{R}^M$ as
\begin{IEEEeqnarray}{rCl}
\label{eq_vector_alpha}  \alpha_j(\beta^{-1}) & := & \left( K_{\mathrm{x}\mathrm{a}}^T K_{\mathrm{x}\mathrm{a}}   + \tau  K_{\mathrm{a}\mathrm{a}}   +    \beta^{-1} K_{\mathrm{a}\mathrm{a}} \right)^{-1}  (K_{\mathrm{x}\mathrm{a}} )^T \mathrm{y}_j  \IEEEeqnarraynumspace
  \end{IEEEeqnarray}
so that $\widehat{ \mathcal{F}_j(x^{i})}$ could be expressed as    
\begin{IEEEeqnarray}{rCl}
\label{eq_final_layer_output} \widehat{ \mathcal{F}_j(x^{i})} & = & \left(G(x^i) \right) \alpha_j(\beta^{-1}).
  \end{IEEEeqnarray} 
It follows from (\ref{eq_final_layer_output}) that estimation of the membership-mapping outputs requires computing $\alpha_j$ via (\ref{eq_vector_alpha}) which in-turn requires estimating the inverse precision value $\beta^{-1}$. The inverse precision value $\beta^{-1}$ is iteratively estimated as the inverse of the mean squared error between data and membership-mappings outputs. That is, 
\begin{IEEEeqnarray}{rCl}
\label{eq_738497.4473} \beta^{-1} & = & \frac{1}{pN}\sum_{j=1}^p \sum_{i=1}^N \left |y_j^i -  \widehat{ \mathcal{F}_j(x^{i})} \right |^2
 \end{IEEEeqnarray} 
where $\widehat{ \mathcal{F}_j(x^{i})}$ is the estimated membership-mapping output given as in~(\ref{eq_final_layer_output}). Expression (\ref{eq_738497.4473}) using (\ref{eq_final_layer_output}) can be expressed as 
\begin{IEEEeqnarray}{rCl}
\label{eq_738567.4264}  \beta^{-1} & = & \frac{1}{pN}\sum_{j=1}^p \sum_{i=1}^N \left |y_j^i -  \left(G(x^i) \right) \alpha_j(\beta^{-1}) \right |^2.
\end{IEEEeqnarray}        
As $G(x^i)$ is equal to the $i-$th row of matrix $K_{\mathrm{x}\mathrm{a}}$, (\ref{eq_738567.4264}) can be expressed as
\begin{IEEEeqnarray}{rCl}
\label{eq_738570.4144} \beta^{-1} & = & \frac{1}{pN}\sum_{j=1}^p \| \mathrm{y}_j  - K_{\mathrm{x}\mathrm{a}} \alpha_j(\beta^{-1}) \|^2.
  \end{IEEEeqnarray}
We suggest to estimate $\beta^{-1}$ and $\alpha_j$ iteratively using (\ref{eq_738570.4144}) and (\ref{eq_vector_alpha}) till the convergence.  
\section{A Globally Convergent Learning Algorithm and Robustness Analysis for Variational Membership-Mappings}\label{sec_convergence_robustness}
\subsection{Convergence Analysis}
In this subsection, we study the convergence of estimation algorithm~(\ref{eq_738570.4144}, \ref{eq_vector_alpha}). In particular, we derive a sufficient condition for estimation algorithm~(\ref{eq_738570.4144}, \ref{eq_vector_alpha}) to converge. For this, consider the singular value decomposition of $K_{\mathrm{x}\mathrm{a}} $:
\begin{IEEEeqnarray}{rCl}
\label{eq_738576.4462}K_{\mathrm{x}\mathrm{a}}  & = & U \left[ \begin{array}{c} S \\ 0 \end{array}\right] V^T
 \end{IEEEeqnarray}    
where $U \in \mathbb{R}^{N \times N}$ and $V \in \mathbb{R}^{M \times M}$ are orthogonal, and $S = \text{diag}(s_1,\cdots,s_M)$ is a diagonal matrix with $s_1 \geq s_2 \geq \cdots \geq s_M \geq 0$ being the singular values of $K_{\mathrm{x}\mathrm{a}} $. The vectors $b_j^1 \in \mathbb{R}^M$ and $b_j^2 \in \mathbb{R}^{N-M}$ are defined as 
\begin{IEEEeqnarray}{rCl}
\label{eq_738574.5863}\left[ \begin{array}{c} b_j^1 \\ b_j^2 \end{array}\right] & = & U^T \mathrm{y}_j .
 \end{IEEEeqnarray}
The expression (\ref{eq_vector_alpha}) for $\alpha_j $ can be rewritten as
\begin{IEEEeqnarray}{rCl}
 \alpha_j(\beta^{-1}) & = & \left( V S^2 V^T   + \tau  K_{\mathrm{a}\mathrm{a}}   +  \beta^{-1}  K_{\mathrm{a}\mathrm{a}} \right)^{-1}  V S b_j^1 .
  \end{IEEEeqnarray}
Consider
\begin{IEEEeqnarray}{rCl}
\lefteqn{\mathrm{y}_j  - K_{\mathrm{x}\mathrm{a}} \alpha_j(\beta^{-1})  =   U U^T \mathrm{y}_j}\\
\nonumber && {-}\: U \left[ \begin{array}{c} S \\ 0 \end{array}\right] V^T \left( V S^2 V^T   + (\tau   +  \beta^{-1} ) K_{\mathrm{a}\mathrm{a}} \right)^{-1}  V S b_j^1. \IEEEeqnarraynumspace 
  \end{IEEEeqnarray}
Using matrix inversion lemma,
\begin{IEEEeqnarray}{rCl}
\lefteqn{\mathrm{y}_j  - K_{\mathrm{x}\mathrm{a}} \alpha_j(\beta^{-1})=}\\
\nonumber &  & U \left[\begin{array}{c} \left( I + \frac{1}{\tau + \beta^{-1}} S V^T  K_{\mathrm{a}\mathrm{a}}^{-1} VS   \right)^{-1}  b_j^1 \\
 b_j^2  \end{array}\right ], \IEEEeqnarraynumspace 
  \end{IEEEeqnarray}
and hence,
\begin{IEEEeqnarray}{rCl}
\label{eq_738569.5471} \lefteqn{ \| \mathrm{y}_j  - K_{\mathrm{x}\mathrm{a}} \alpha_j(\beta^{-1})\|^2  =  \|  b_j^2 \|^2} \\
\nonumber && {+}\: (\tau + \beta^{-1})^2(b_j^1)^T \left( (\tau + \beta^{-1})I + S V^T  K_{\mathrm{a}\mathrm{a}}^{-1} VS   \right)^{-2}  b_j^1
  \end{IEEEeqnarray}
Using (\ref{eq_738569.5471}) in (\ref{eq_738570.4144}),
\begin{IEEEeqnarray}{rCl}
\label{eq_738569.5967} \lefteqn{\beta^{-1}   =  \frac{1}{pN}\sum_{j=1}^p  \|  b_j^2 \|^2} \\
\nonumber && {+}\: \frac{ (\tau +\beta^{-1})^2}{pN} \sum_{j=1}^p (b_j^1)^T \left( (\tau + \beta^{-1})I + S V^T  K_{\mathrm{a}\mathrm{a}}^{-1} VS   \right)^{-2}  b_j^1.
  \end{IEEEeqnarray}
\begin{result}[A Function in $\beta^{-1}$]\label{result_definition_variance_function}
Let $\mathcal{R}$ be a function in $\beta^{-1}$ defined as
\begin{IEEEeqnarray}{rCl}
\label{eq_738570.8644} \lefteqn{\mathcal{R}(\beta^{-1})  : =  \frac{1}{pN}\sum_{j=1}^p  \|  b_j^2 \|^2} \\
\nonumber && + \frac{(\tau + \beta^{-1})^2}{pN}\sum_{j=1}^p  (b_j^1)^T \left( (\tau + \beta^{-1})I + S V^T  K_{\mathrm{a}\mathrm{a}}^{-1} VS   \right)^{-2}  b_j^1.
  \end{IEEEeqnarray}
We have followings:
\begin{enumerate}
\item $\mathcal{R}(\beta^{-1}) \in \left(\beta^{-1}|_{low}, \beta^{-1}|_{up}  \right), \; \forall \beta^{-1} \in \mathbb{R}_{>0}$, where 
\begin{IEEEeqnarray}{rCl}
\label{eq_738576.4643} \beta^{-1}|_{low} &  = & \frac{1}{pN}\sum_{j=1}^p  \|  b_j^2 \|^2 \\
\label{eq_738576.4646} \beta^{-1}|_{up} &  = &  \frac{1}{pN}\sum_{j=1}^p  \| \mathrm{y}_j  \|^2.
  \end{IEEEeqnarray}    
\item The lower and upper bounds on the derivative of $\mathcal{R}$ w.r.t. $\beta^{-1}$ are given as
\begin{align}
0  <  \frac{\dd \mathcal{R}(\beta^{-1})}{\dd \beta^{-1}}  <  \frac{2}{pN} \frac{1}{\tau +\beta^{-1}} \sum_{j=1}^p  \|  b_j^1 \|^2. \label{eq_738573.4562}
\end{align}   
\item $\mathcal{R}(\beta^{-1})$ has at least one fixed point in $\left(\beta^{-1}|_{low}, \beta^{-1}|_{up}  \right)$.
\end{enumerate}
\end{result} 
\begin{IEEEproof}
The proof is provided in Appendix \ref{appendix_738612.5557}.
\end{IEEEproof}
\begin{result}[Convergence]\label{result_convergence}
If $\tau$ is chosen such that
\begin{IEEEeqnarray}{rCl}
\label{eq_738575.721}\tau & > & \frac{2}{pN}\sum_{j=1}^p  \| \mathrm{y}_j  \|^2 - \frac{1}{pN}\sum_{j=1}^p  \|  b_j^2 \|^2,
  \end{IEEEeqnarray} 
then the iterations
\begin{IEEEeqnarray}{rCl}
\label{eq_738573.4879}\beta^{-1}|_{it+1} & = & \mathcal{R}(\beta^{-1}|_{it}),\;it \in \{0,1,2,\cdots \},
  \end{IEEEeqnarray} 
with $\beta^{-1}|_{0} \in \left(\beta^{-1}|_{low}, \beta^{-1}|_{up}  \right)$, converge to the unique fixed point of $\mathcal{R}(\beta^{-1})$.
\end{result}
\begin{IEEEproof}
The proof is provided in Appendix \ref{appendix_738612.6484}. 
\end{IEEEproof}
\subsection{Learning Algorithm}
Result~\ref{result_convergence} allows to design a globally convergent algorithm for the variational learning of membership-mappings via ensuring the sufficient condition~(\ref{eq_738575.721}). For this, it is observed that $\tau$, defined as in (\ref{eq_738566.838771}), is a function of parameters set $\{M, \sigma^2,\{w_i\}_{i=1}^n,\mathrm{x},\mathrm{a},\nu\}$. It follows from the kernel function definition~(\ref{eq_membership1003_3}) that
\begin{IEEEeqnarray}{rCl}
\label{eq_738497.3805} \tau(M, \sigma^2,\{w_i\}_{i=1}^n,\mathrm{x},\mathrm{a},\nu) & = & \sigma^2\tau(M, 1,\{w_i\}_{i=1}^n,\mathrm{x},\mathrm{a},\nu).  \IEEEeqnarraynumspace
 \end{IEEEeqnarray}    
Using (\ref{eq_738497.3805}), the condition (\ref{eq_738575.721}) can be rewritten as
\begin{IEEEeqnarray}{rCl}
 \label{eq_738497.3839} \sigma^2 & > & \frac{1}{\tau(M, 1,\{w_i\}_{i=1}^n,\mathrm{x},\mathrm{a},\nu)} \frac{\sum_{j=1}^p(2\| \mathrm{y}_j  \|^2- \|  b_j^2 \|^2)}{pN}. \IEEEeqnarraynumspace
 \end{IEEEeqnarray}  
To hold the inequality (\ref{eq_738497.3839}), the value of $\sigma^2$ is adjusted as in the following:
\begin{algorithmic}
\IF{$\tau(M, 1,\{w_i\}_{i=1}^n,\mathrm{x},\mathrm{a},\nu) > \frac{\sum_{j=1}^p(2\| \mathrm{y}_j  \|^2- \|  b_j^2 \|^2)}{pN}$}
\STATE $\sigma^2 = 1$
\ELSE 
\STATE $\sigma^2 =  \frac{1.1 }{\tau(M, 1,\{w_i\}_{i=1}^n,\mathrm{x},\mathrm{a},\nu)} \frac{\sum_{j=1}^p(2\| \mathrm{y}_j  \|^2- \|  b_j^2 \|^2)}{pN} $.
\ENDIF
\end{algorithmic}
Further, some of the parameters must be prior chosen which are suggested to be chosen as in the following:
\paragraph{Auxiliary inducing points} 
The auxiliary inducing points are suggested to be chosen as the cluster centroids: 
\begin{IEEEeqnarray}{rCl}
\label{eq_738496.4692}\mathrm{a} = \{ a^{m}\}_{m=1}^M  =  cluster\_centroid(  \{x^i\}_{i=1}^N, M) 
 \end{IEEEeqnarray} 
where $cluster\_centroid(  \{ x^i \}_{i=1}^N,M)$ represents the k-means clustering on $ \{ x^i \}_{i=1}^N$.  
\paragraph{Degrees of freedom}
The degrees of freedom associated to the Student-t membership-mapping $\nu \in \mathbb{R}_{+} \setminus [0,2]$ is chosen as 
\begin{IEEEeqnarray}{rCl}
\label{eq_738496.4701}\nu & = & 2.1
 \end{IEEEeqnarray} 
\paragraph{Parameters $(w_1,\cdots,w_n)$}
The parameters $(w_1,\cdots,w_n)$ for kernel function~(\ref{eq_membership1003_3}) are chosen such that $w_{k}$ (for $k\in \{1,2,\cdots,n\}$) is given as
\begin{IEEEeqnarray}{rCl}
\label{eq_738496.4698}w_k & = & \left(\max_{1 \leq i \leq N}\left(x^i_k\right) - \min_{1 \leq i \leq N}\left(x^i_k\right)\right)^{-2}
 \end{IEEEeqnarray} 
where $x^i_k$ is the $k-$th element of vector $x^i \in \mathbb{R}^n$.

Finally, Algorithm~\ref{algorithm_convergent_basic_learning} presents a systematic procedure for the variational learning of membership-mappings while ensuring the sufficient condition~(\ref{eq_738575.721}) for the convergence.   
\begin{algorithm}
\caption{A globally convergent algorithm for the variational learning of membership-mappings}
\label{algorithm_convergent_basic_learning}
{\footnotesize
\begin{algorithmic}[1]
\REQUIRE  Dataset $\left\{ (x^i,y^i) \; | \;  i \in \{1,\cdots,N \} \right \}$ and maximum possible number of auxiliary points $M_{max} \in \mathbb{Z}_+$ with $M_{max} \leq N$.  
\STATE Choose $\nu$ and $w = (w_1,\cdots,w_n)$ as in (\ref{eq_738496.4701}) and (\ref{eq_738496.4698}) respectively.  
\STATE Set iteration count $it = 0$, $M|_0 = M_{max}$, and determine $\mathrm{a}|_0 = \{ a^{m}|_0\}_{m=1}^{M|_0}$ using (\ref{eq_738496.4692}).
\WHILE{$\tau(M|_{it},1,\{w_i\}_{i=1}^n,\mathrm{x},\mathrm{a}|_{it},\nu) \leq 0$}
\STATE $M|_{it+1} =  M|_{it} - 1$
\STATE Determine $\mathrm{a}|_{it+1} = \{ a^{m}|_{it+1}\}_{m=1}^{M|_{it+1}}$ using (\ref{eq_738496.4692}).
\STATE $it \leftarrow it + 1$
\ENDWHILE
\STATE Set $M = M|_{it}$ and compute $\mathrm{a} = \{ a^{m}\}_{m=1}^M$ using (\ref{eq_738496.4692}).
\STATE Compute $K_{\mathrm{x}\mathrm{a}}$ using (\ref{eq_738497.4922}) taking $\sigma^2 = 1$ and perform singular value decomposition of $K_{\mathrm{x}\mathrm{a}}$ to compute orthogonal matrix $U$ such that (\ref{eq_738576.4462}) holds. Further compute $b_j^1$ and $b_j^2$ using (\ref{eq_738574.5863}) for all $j \in \{1,\cdots,p\}$.
\IF{$\tau(M, 1,\{w_i\}_{i=1}^n,\mathrm{x},\mathrm{a},\nu) > \frac{\sum_{j=1}^p(2\| \mathrm{y}_j  \|^2- \|  b_j^2 \|^2)}{pN}$}
\STATE $\sigma^2 = 1$
\ELSE 
\STATE $\sigma^2 =  \frac{1.1 }{\tau(M, 1,\{w_i\}_{i=1}^n,\mathrm{x},\mathrm{a},\nu)} \frac{\sum_{j=1}^p(2\| \mathrm{y}_j  \|^2- \|  b_j^2 \|^2)}{pN} $.
\ENDIF
\STATE Compute $\mathrm{a} = \{ a^{m}\}_{m=1}^M$ using (\ref{eq_738496.4692}), $K_{\mathrm{x}\mathrm{x}}$ using (\ref{738026.844153}), $K_{\mathrm{a}\mathrm{a}}$ using (\ref{eq_membership1004_2}), and $K_{\mathrm{x}\mathrm{a}}$ using (\ref{eq_738497.4922}).
\STATE Compute $\tau$ using (\ref{eq_738566.838771}).
\STATE Set iteration count $it = 0$ and $\beta^{-1}|_0 = 0.5(\beta^{-1}|_{low} + \beta^{-1}|_{up})$, where $\beta^{-1}|_{low}$ and $\beta^{-1}|_{up}$ are given by (\ref{eq_738576.4643}) and (\ref{eq_738576.4646}) respectively.
\STATE Determine the unique fixed point of $\mathcal{R}(\beta^{-1})$, say $\hat{\beta}^{-1}$, using iterations~(\ref{eq_738573.4879}). 
\STATE Compute matrix $\alpha = \left[\begin{IEEEeqnarraybox*}[][c]{,c/c/c,}  \alpha_1(\hat{\beta}^{-1}) & \cdots & \alpha_p(\hat{\beta}^{-1})
 \end{IEEEeqnarraybox*} \right] \in \mathbb{R}^{M \times p}$, where $\alpha_j(\hat{\beta}^{-1})$, $j\in \{1,\cdots,p\}$, is computed using (\ref{eq_vector_alpha}).
\RETURN the parameters set $\mathbb{M} = \{\alpha, \mathrm{a}, M,\sigma,w,\hat{\beta}\}$.
\end{algorithmic} 
}
\end{algorithm}
\begin{definition}[Membership-Mappings Prediction]
Given the parameters set $\mathbb{M} = \{\alpha, \mathrm{a}, M,\sigma,w,\hat{\beta}\}$ returned by Algorithm~\ref{algorithm_convergent_basic_learning}, the learned membership-mappings could be used to predict output corresponding to any arbitrary input data point $x \in \mathbb{R}^n$ as
\begin{IEEEeqnarray}{rCl}
\hat{y}(x;\mathbb{M}) & = & \left[\begin{IEEEeqnarraybox*}[][c]{,c/c/c,} \widehat{ \mathcal{F}_1(x)} & \cdots & \widehat{ \mathcal{F}_p(x)}\end{IEEEeqnarraybox*} \right]^T
 \end{IEEEeqnarray}  
where $\widehat{ \mathcal{F}_j(x)}$, defined as in (\ref{eq_final_layer_output}), is the estimated output of $j-$th membership-mapping. It follows from (\ref{eq_final_layer_output}) that  
\begin{IEEEeqnarray}{rCl}
\label{eq_738124.770095}\hat{y}(x;\mathbb{M}) & = & (\alpha(\hat{\beta}))^T(G(x))^T
\end{IEEEeqnarray}
where $G(\cdot) \in \mathbb{R}^{1 \times M}$ is a vector-valued function~(\ref{eq_738495.5497}). 
\end{definition}
As a CDMMA consists of membership-mapping compositions, Algorithm~\ref{algorithm_convergent_basic_learning} can be directly applied for their learning as in Algorithm~\ref{algorithm_DSFMA} and Algorithm~\ref{algorithm_optimal_RWDSFMA}. For practical applications, Algorithm~\ref{algorithm_optimal_RWDSFMA} is suggested for the learning of wide CDMMA where a computational optimization of a free parameter is performed via minimizing the estimated variance of the mean squared error between data and membership-mappings outputs. 
\begin{algorithm}
\caption{A globally convergent algorithm for the variational learning of CDMMA}
\label{algorithm_DSFMA}
{\footnotesize
\begin{algorithmic}[1]
\REQUIRE Data set $\mathbf{Y} = \left\{ y^i \in \mathbb{R}^p \; | \; i \in \{1,\cdots,N \} \right \}$; the subspace dimension $n \in \{1,2,\cdots,p \}$; maximum number of auxiliary points $M_{max} \in \mathbb{Z}_+$ with $M_{max} \leq N$; the number of layers $L \in \mathbb{Z}_{+}$.
\FOR{$l=1$ to $L$}
\STATE Set subspace dimension associated to $l-$th layer as $n_l = \max(n - l + 1,1)$.
\STATE Define $P^l \in \mathbb{R}^{n_l \times p}$ such that $i-$th row of $P^l$ is equal to transpose of eigenvector corresponding to $i-$th largest eigenvalue of sample covariance matrix of data set $\mathbf{Y} $. 
\STATE Define a latent variable $x^{l,i} \in \mathbb{R}^{n_l}$, for $i \in \{1,\cdots,N \}$, as
 \begin{IEEEeqnarray}{rCl}
\label{eq_x_l_i}x^{l,i} &:=& \left\{ \,
    \begin{IEEEeqnarraybox}[][c]{l?s}
      \IEEEstrut
      P^ly^i & if $l=1$, \\
     P^l \hat{y}^{l-1}(x^{l-1,i};\mathbb{M}^{l-1}) & if $l > 1$
      \IEEEstrut
    \end{IEEEeqnarraybox}
\right.  \IEEEeqnarraynumspace
\end{IEEEeqnarray}   
where $\hat{y}^{l-1}$ is the estimated output of the $(l-1)-$th layer computed using (\ref{eq_738124.770095}) for the parameters set $\mathbb{M}^{l-1} = \{\alpha^{l-1}, \mathrm{a}^{l-1}, M^{l-1}, \sigma^{l-1},w^{l-1}\}$.  
\STATE Define $M_{max}^l$ as
 \begin{IEEEeqnarray}{rCl}
\label{eq_738499.4927}M_{max}^l &:=& \left\{ \,
    \begin{IEEEeqnarraybox}[][c]{l?s}
      \IEEEstrut
      M_{max} & if $l=1$, \\
     M^{l-1} & if $l > 1$
      \IEEEstrut
    \end{IEEEeqnarraybox}
\right.  \IEEEeqnarraynumspace
\end{IEEEeqnarray} 
\STATE Compute parameters set $\mathbb{M}^l = \{\alpha^{l}, \mathrm{a}^{l}, M^{l}, \sigma^{l},w^{l}, \hat{\beta}^{l}\}$, characterizing the membership-mappings associated to $l-$th layer, using Algorithm~\ref{algorithm_convergent_basic_learning} on data set $\left\{ (x^{l,i},y^i) \; | \;  i \in \{1,\cdots,N \} \right \}$ with maximum possible number of auxiliary points $M_{max}^l$. 
\ENDFOR
\RETURN the parameters set $\mathcal{M} = \{\{\mathbb{M}^1,\cdots,\mathbb{M}^L\}, \{P^1,\cdots,P^L \} \}$.
\end{algorithmic}
}
\end{algorithm} 
\begin{definition}[CDMMA Filtering]\label{def_DSFMA_filtering}
Given a CDMMA with its parameters being represented by a set $\mathcal{M} = \{\{\mathbb{M}^1,\cdots,\mathbb{M}^L\}, \{P^1,\cdots,P^L \} \}$, the autoencoder can be applied for filtering a given input vector $y \in \mathbb{R}^p$ as follows:   
 \begin{IEEEeqnarray}{rCl}
x^l(y;\mathcal{M}) &=& \left\{ \,
    \begin{IEEEeqnarraybox}[][c]{l?s}
      \IEEEstrut
      P^ly, & $l=1$ \\
      P^l  \hat{y}^{l-1}(x^{l-1};\mathbb{M}^{l-1})&  $l \geq 2$
      \IEEEstrut
    \end{IEEEeqnarraybox}
\right. 
\end{IEEEeqnarray} 
Here, $\hat{y}^{l-1}$ is the output of the $(l-1)-$th layer estimated using (\ref{eq_738124.770095}). Finally, CDMMA's output, $\mathcal{D}(y;\mathcal{M})$, is given as
\begin{IEEEeqnarray}{rCl}
\label{eq_satguru_18}  \widehat{\mathcal{D}}(y;\mathcal{M}) & = &  \hat{y}^{l^*}(x^{l^*};\mathbb{M}^{l^*}) \\
\label{eq_satguru_19} l^*  & = & \arg\;\min_{l \: {\in} \: \{1,\cdots,L \}}\; \|y - \hat{y}^{l}(x^{l};\mathbb{M}^{l}) \|^2.
 \end{IEEEeqnarray} 
\end{definition}
\begin{algorithm}
\caption{A globally convergent algorithm for the variational learning of wide CDMMA}
\label{algorithm_optimal_RWDSFMA}
{\footnotesize
\begin{algorithmic}[1]
\REQUIRE  Data set $\mathbf{Y} = \left\{ y^i \in \mathbb{R}^p \; | \; i \in \{1,\cdots,N \} \right \}$; the number of layers $L \in \mathbb{Z}_{+}$; the subspace dimension $n \in \{1,2,\cdots,p\}$; an array of possible $r_{max}$ values $\{r_{max}^1,\cdots,r_{max}^{N_r}\}$ with $0< r_{max}^1 < r_{max}^2 < \cdots < r_{max}^{N_r} \leq 1$.
\STATE Apply k-means clustering to partition $\mathbf{Y} $ into $S$ subsets, $\{\mathbf{Y}^1, \cdots, \mathbf{Y}^S  \}$, where $S = \lceil N/1000 \rceil$. 
\FOR{$s = 1$ to $S$}
\FOR{$r = r_{max}^1$ to $r = r_{max}^{N_r}$}
\STATE Apply Algorithm~\ref{algorithm_DSFMA} on $\mathbf{Y}^s$ to build a single-layered CDMMA, $\mathcal{M}^{r} = \{\{\mathbb{M}^{1,r}\}, \{P^{1,r} \} \}$ where $\mathbb{M}^{1,r} = \{\alpha^{1,r}, \mathrm{a}^{1,r}, M^{1,r}, \sigma^{1,r},w^{1,r}, \hat{\beta}^{1,r}\}$, taking $n$ as the subspace dimension; maximum number of auxiliary points as equal to $r \times \#\mathbf{Y}^s$ (where $\#\mathbf{Y}^s$ is the number of data points in $\mathbf{Y}^s$); and $L = 1$.  
\ENDFOR
\STATE Set $r_{max} = \arg \; \max_{r \in \{r_{max}^1, r_{max}^2, \cdots, r_{max}^{N_r}\}} \hat{\beta}^{1,r}$.
\STATE Build a CDMMA, $\mathcal{M}^s$, by applying Algorithm~\ref{algorithm_DSFMA} on $\mathbf{Y}^s$ taking $n$ as the subspace dimension; maximum number of auxiliary points as equal to $r_{max} \times \#\mathbf{Y}^s$ (where $\#\mathbf{Y}^s$ is the number of data points in $\mathbf{Y}^s$); and $L$ as the number of layers.   
\ENDFOR
\RETURN the parameters set $\mathcal{P} = \{\mathcal{M}^s\}_{s=1}^S$.
\end{algorithmic}
}
\end{algorithm}
\begin{definition}[Wide CDMMA Filtering]\label{def_WDSFMA_filtering}
Given a wide CDMMA with its parameters being represented by a set $\mathcal{P} = \{\mathcal{M}^s\}_{s=1}^S$, the autoencoder can be applied for filtering a given input vector $y \in \mathbb{R}^p$ as follows:   
  \begin{IEEEeqnarray}{rCl}
 \label{eq_738500.4495}\widehat{\mathcal{WD}}(y;\mathcal{P}) & = &   \widehat{\mathcal{D}}(y;\mathcal{M}^{s^*})\\
 s^* & = & \arg\;\min_{s \in \{1,2,\cdots,S \}}\; \| y -  \widehat{\mathcal{D}}(y;\mathcal{M}^{s})  \|^2,
 \end{IEEEeqnarray}   
where $ \widehat{\mathcal{D}}(y;\mathcal{M}^{s})$ is the output of $s-$th CDMMA estimated using (\ref{eq_satguru_18}).
\end{definition}
\subsection{Robustness Analysis}
Consider that the data samples $\{ (x^i,y^i) \; | \; i \in \{1,\cdots,N\}\}$ are subject to deterministic perturbations and thus matrix $K_{\mathrm{x}\mathrm{a}}$ (\ref{eq_738497.4922}) and vector $\mathrm{y}_j$ (\ref{eq_y_j_vec2000}) are subject to perturbations. Let $\Delta K_{\mathrm{x}\mathrm{a}} \in \mathbb{R}^{N \times M}$ and $\Delta \mathrm{y}_j \in \mathbb{R}^{N}$ be the unknown (but bounded) perturbations in $K_{\mathrm{x}\mathrm{a}}$ and $\mathrm{y}_j$ respectively due to the perturbations in the data samples. The data model assumes that there exists some parameters vector $\alpha^*_j \in \mathbb{R}^M$ such that
 \begin{IEEEeqnarray}{rCl}  
\label{eq_738579.5214} \mathrm{y}_j + \Delta \mathrm{y}_j & = & (K_{\mathrm{x}\mathrm{a}} + \Delta K_{\mathrm{x}\mathrm{a}})\alpha_j^*.
 \end{IEEEeqnarray}   
The modeling problem is concerned with the estimation of $\alpha_j^*$ in the presence of unknown perturbations $\Delta K_{\mathrm{x}\mathrm{a}}$ and $\Delta \mathrm{y}_j $. To show that Algorithm~\ref{algorithm_convergent_basic_learning} provides a robust estimation of $\alpha_j^*$, define
 \begin{IEEEeqnarray}{rCl}  
\Delta_x & := & \Delta K_{\mathrm{x}\mathrm{a}} (K_{\mathrm{a}\mathrm{a}}^{1/2})^{-1}  
 \end{IEEEeqnarray}             
where $K_{\mathrm{a}\mathrm{a}}^{1/2}$ is the unique square root of positive definite matrix $K_{\mathrm{a}\mathrm{a}} $ (\ref{eq_membership1004_2}). The set of equations~(\ref{eq_738579.5214}) can be expressed as
 \begin{IEEEeqnarray}{rCl}  
 \mathrm{y}_j + \Delta \mathrm{y}_j & = & (K_{\mathrm{x}\mathrm{a}} + \Delta_x K_{\mathrm{a}\mathrm{a}}^{1/2})\alpha_j^*.
 \end{IEEEeqnarray}   
The perturbation matrix $\left[\begin{IEEEeqnarraybox*}[][c]{,c/c,} \Delta_x &  \Delta \mathrm{y}_j\end{IEEEeqnarraybox*} \right]$ is unknown, however, is assumed bounded. That is, there exists a scalar $\delta_{m} > 0$ such that $\| \left[\begin{IEEEeqnarraybox*}[][c]{,c/c,} \Delta_x &  \Delta \mathrm{y}_j\end{IEEEeqnarraybox*} \right] \|_F \leq \delta_{m} $, where $\| \cdot \|_F$ denotes the Frobenius norm. A robust solution to the estimation of parameters seeks to alleviate the worst-case effect of perturbations. For example, the worst-case residual error can be minimized via solving a min-max estimation problem as in Result~\ref{result_738582.4835}.         
\begin{result}[Robustness]\label{result_738582.4835}
Algorithm~\ref{algorithm_convergent_basic_learning} provides a robust estimation of $\alpha_j^*$ via solving the following min-max estimation problem:
 \begin{IEEEeqnarray}{rCl}  
\label{eq_738582.4785} \lefteqn{\hat{\alpha}_j  = \arg \; \min_{\alpha_j^*}}\\
\nonumber &&  \max_{\left \| \left[\begin{IEEEeqnarraybox*}[][c]{,c/c,} \Delta_x &  \Delta \mathrm{y}_j\end{IEEEeqnarraybox*} \right] \right \|_F \leq \delta_{m}} \; \left \| (K_{\mathrm{x}\mathrm{a}} + \Delta_x K_{\mathrm{a}\mathrm{a}}^{1/2})\alpha_j^*  - ( \mathrm{y}_j + \Delta \mathrm{y}_j ) \right \|,
 \end{IEEEeqnarray}   
where the upper-bound on the norm of perturbation matrix is given as
 \begin{IEEEeqnarray}{rCl}  
\lefteqn{\delta_{m}  = } \\
\nonumber && \frac{\sqrt{1+ \| K_{\mathrm{a}\mathrm{a}}^{1/2} ( K_{\mathrm{x}\mathrm{a}}^T K_{\mathrm{x}\mathrm{a}}   + \tau  K_{\mathrm{a}\mathrm{a}}   +    \hat{\beta}^{-1} K_{\mathrm{a}\mathrm{a}}  )^{-1}  K_{\mathrm{x}\mathrm{a}}^T \mathrm{y}_j  \|^2}}{\| ( (\tau + \hat{\beta}^{-1})I + K_{\mathrm{x}\mathrm{a}} K_{\mathrm{a}\mathrm{a}}^{-1}K_{\mathrm{x}\mathrm{a}}^T   )^{-1} \mathrm{y}_j   \|}, 
 \end{IEEEeqnarray}     
where $ \hat{\beta}^{-1}$ is the unique fixed point of $\mathcal{R}(\beta^{-1})$ to which the iterations (\ref{eq_738573.4879}) converge. 
\end{result}
\begin{IEEEproof}
The proof is provided in Appendix~\ref{appendix_738614.4953}.
\end{IEEEproof}
\section{Secure Distributed Deep Learning}\label{sec_application}
\subsection{Fuzzy Attributes and Classification Applications}
\begin{figure*}
\subfloat[fuzzy attribute $\mathbf{A}_{\mathcal{P}_1}$]{\includegraphics[width=0.25\textwidth]{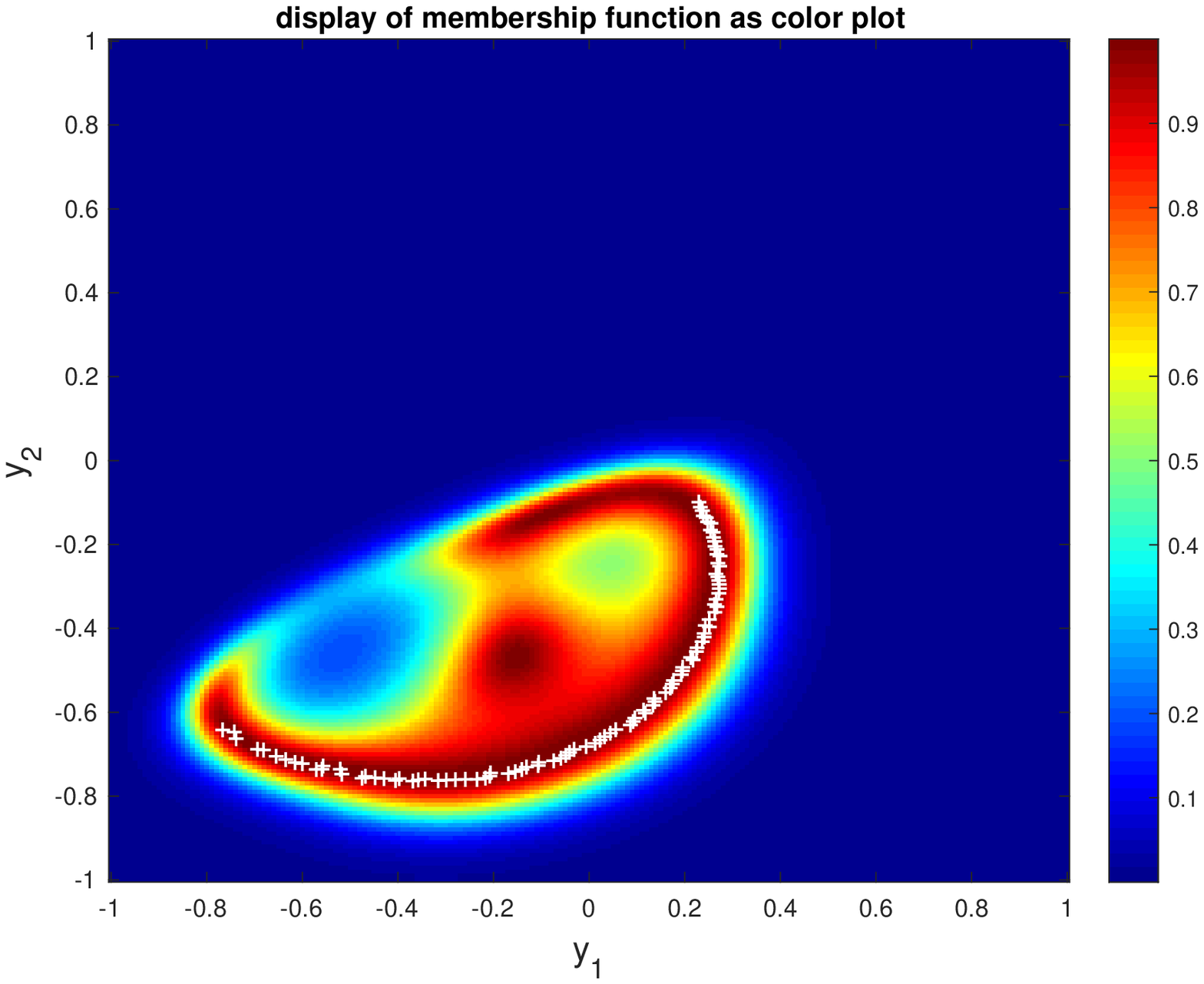}\label{fig_membership_function_1}} \hfil 
\subfloat[fuzzy attribute $\mathbf{A}_{\mathcal{P}_2}$]{\includegraphics[width=0.25\textwidth]{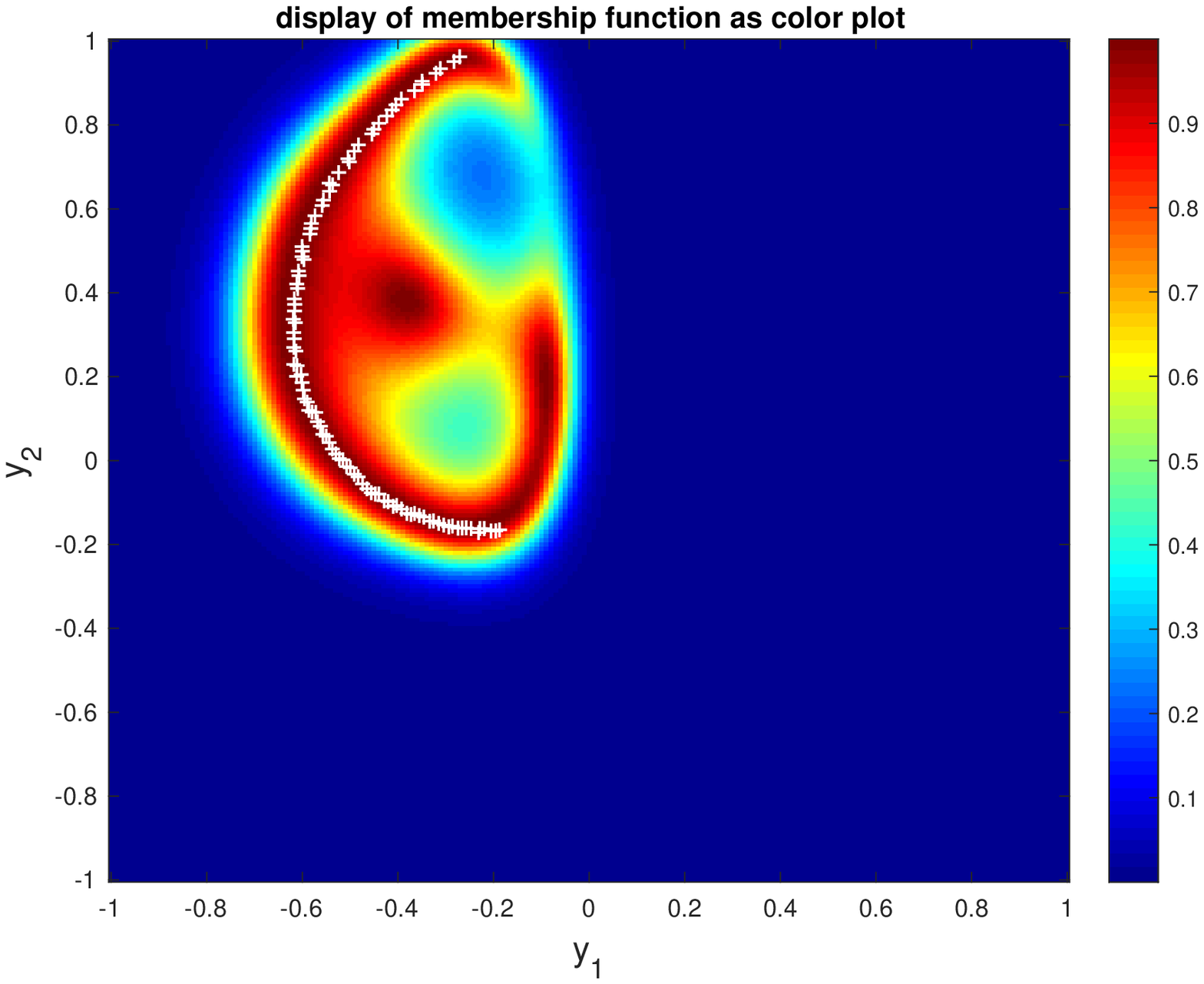}\label{fig_membership_function_2}} \hfil 
\subfloat[fuzzy attribute $\mathbf{A}_{\mathcal{P}_3}$]{\includegraphics[width=0.25\textwidth]{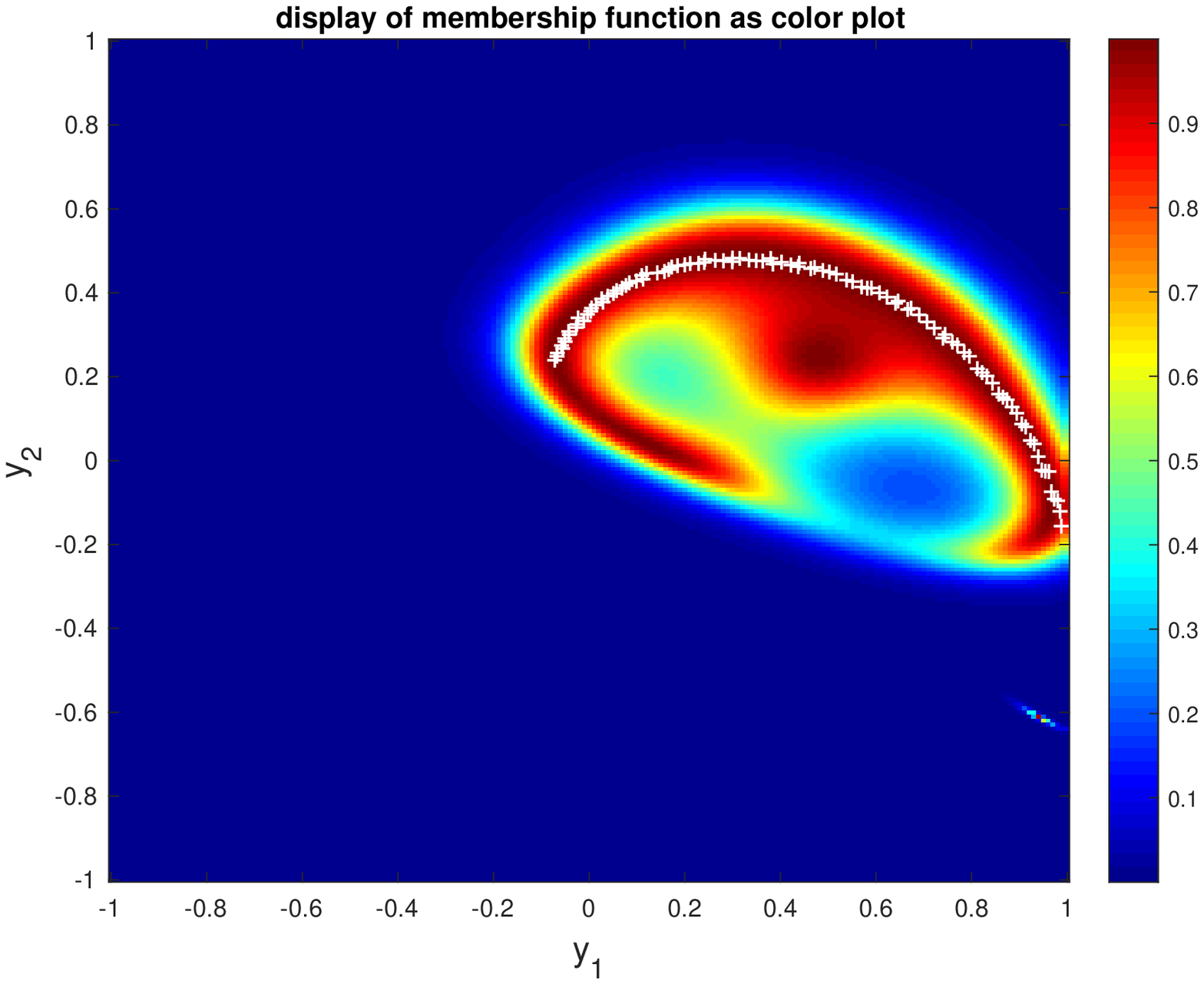}\label{fig_membership_function_3}}  \hfil
\subfloat[decision boundaries determined by the classifier~(\ref{eq_738594.728862})]{\includegraphics[width=0.25\textwidth]{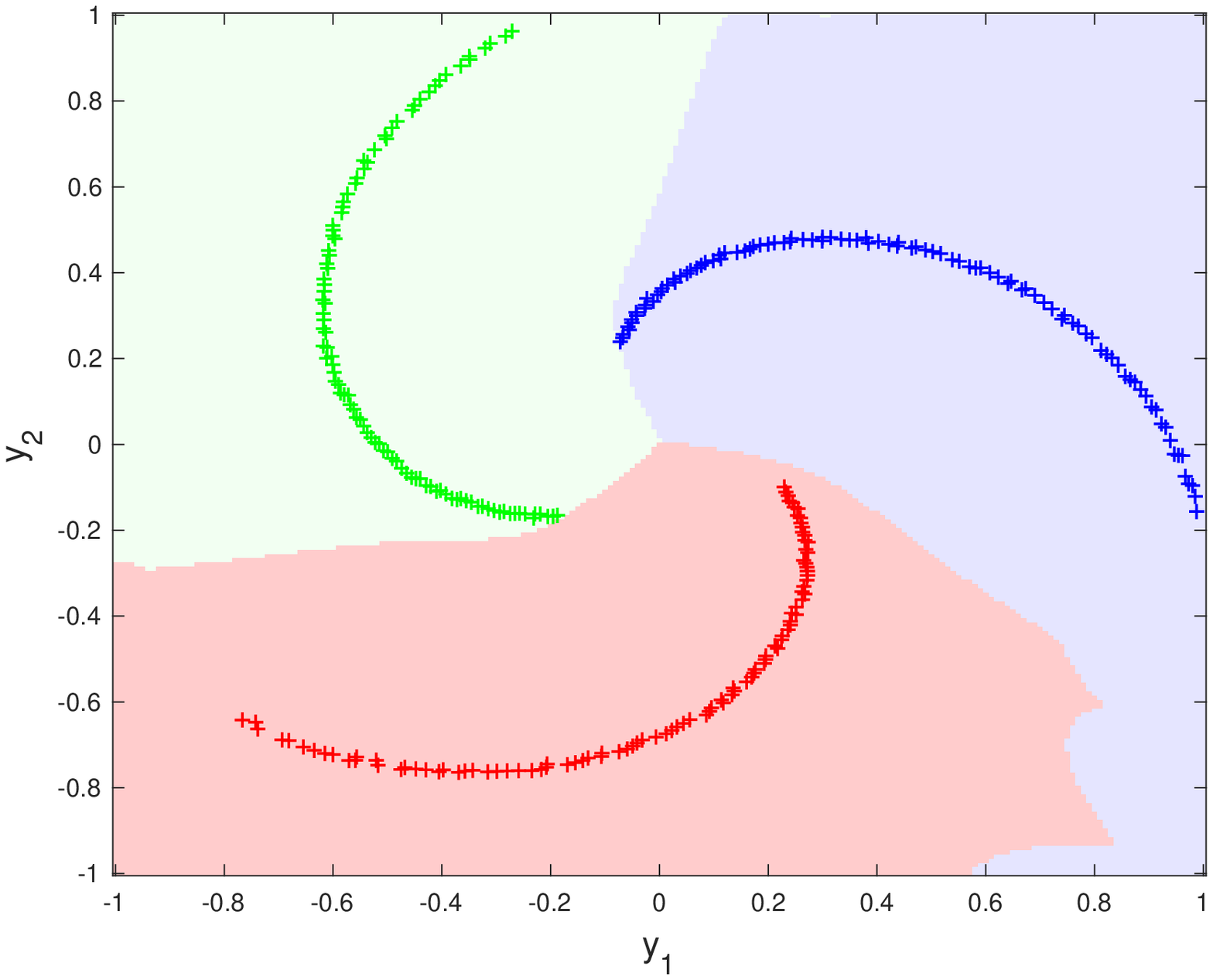}\label{fig_decision_boundaries}}
\caption{A few examples of fuzzy attributes induced by deep autoencoders. The given set of 100 two-dimensional data samples, marked as `+' in the figures, has been used to train a wide conditionally deep membership-mapping autoencoder using Algorithm~\ref{algorithm_optimal_RWDSFMA} taking $L = 5$, $n = 2$, and possible $r_{max}$ values as $\{0.01,0.02,\cdots,0.1\}$. The membership function associated to a fuzzy attribute has been defined by (\ref{eq_738591.853926}) taking $\nu = 2.001$.}
\label{fig_membership_examples}
\end{figure*}
\begin{definition}[A Fuzzy Attribute Induced by Deep Membership-Mapping Autoencoder]\label{def_738589.407817}
A fuzzy attribute $\mathbf{A}_{\mathcal{P}}$, associated to a wide conditionally deep membership-mapping autoencoder $\mathcal{P}$ (that has been learned using dataset $\mathbf{Y} \subset \mathbb{R}^p$), can be defined on a universe of discourse $\mathbb{R}^p$ as 
  \begin{IEEEeqnarray}{rCl}
 \mathbf{A}_{\mathcal{P}} & : = & \left \{ \left(y,\mu_{\mathbf{A}_{\mathcal{P}}}(y)\right)\; | \; y \in \mathbb{R}^p \right \} 
 \end{IEEEeqnarray}  
where $\mu_{\mathbf{A}_{\mathcal{P}}}(y):\mathbb{R}^p \rightarrow [0,1]$ is a $p-$variate membership function such that $\mu_{\mathbf{A}_{\mathcal{P}}}(y)$ is interpreted as the degree to which a point $y \in \mathbb{R}^p$ matches to the attribute $\mathbf{A}_{\mathcal{P}}$. Without the loss of generality, the following Student-t type membership function can be defined to characterize $ \mathbf{A}_{\mathcal{P}} $:
  \begin{IEEEeqnarray}{rCl}
\label{eq_738591.853926}\mu_{\mathbf{A}_{\mathcal{P}}}(y) &  = & \left(1 + \frac{1}{\nu - 2}  \| y - \widehat{\mathcal{WD}}(y;\mathcal{P})\|^2 \right)^{- \frac{\nu + p}{2}},
 \end{IEEEeqnarray}  
where $\widehat{\mathcal{WD}}(y;\mathcal{P})$ is through autoencoder $\mathcal{P}$ filtered output (\ref{eq_738500.4495}). Similarly, one can define Gaussian type of membership function:
  \begin{IEEEeqnarray}{rCl}
  \label{eq_738595.3795} \mu_{\mathbf{A}_{\mathcal{P}}}(y) &  = & \exp\left(- \frac{1}{2p} \| y - \widehat{\mathcal{WD}}(y;\mathcal{P})\|^2  \right).
 \end{IEEEeqnarray}    
\end{definition} 
A wide conditionally deep membership-mapping autoencoder induces a fuzzy attribute as defined in Definition~\ref{def_738589.407817}. Fig.~\ref{fig_membership_examples} provides three different examples of fuzzy attributes induced by the deep autoencoders. As demonstrated through color-plots in Fig.~\ref{fig_membership_examples}, the defined fuzzy attribute (Definition~\ref{def_738589.407817}) learns a representation of the data samples. This motivates to define a fuzzy classifier based on the following if-then rules:
  \begin{IEEEeqnarray}{CCC}
 \nonumber \mbox{If $y$ is $\mathbf{A}_{\mathcal{P}_1}$} &, & \mbox{then the class is 1};\\
 \label{eq_738594.721927} & \vdots & \\
\nonumber  \mbox{If $y$ is $\mathbf{A}_{\mathcal{P}_C}$} &, & \mbox{then the class is C}.
   \end{IEEEeqnarray}       
The class-label associated to a data point $y$ is predicted based on fuzzy rules~(\ref{eq_738594.721927}) as
  \begin{IEEEeqnarray}{rCl}
\label{eq_738594.728862}\mathcal{C}(y;\{ \mathcal{P}_c\}_{c=1}^C) & = & \arg \; \max_{1 \leq c \leq C} \; \mu_{\mathbf{A}_{\mathcal{P}_c}}(y).
   \end{IEEEeqnarray}      
The classifier~(\ref{eq_738594.728862}), $\mathcal{C}:\mathbb{R}^p \rightarrow \{1,2,\cdots,C \}$, assigns to an input vector the label of the class to which the data point has highest degree of matching. An example of the decision boundary determined by a three-class classifier is provided in Fig.~\ref{fig_membership_examples}. 
\subsection{Distributed Learning}
We consider a scenario that data are distributed amongst different parties. Assume that there are $K$ different datasets, $\{\mathbf{Y}^1,\cdots,\mathbf{Y}^K \}$, owned locally by $K$ different parties. We consider the multi-class classification problem assuming that each local dataset, say $\mathbf{Y}^k$, can be partitioned into $C$ different classes, i.e.,
  \begin{IEEEeqnarray}{rCl}
  \mathbf{Y}^k & = & \{ \mathbf{Y}^k_1,\cdots, \mathbf{Y}^k_C \}
   \end{IEEEeqnarray}
where $\mathbf{Y}^k_c$ refers to the $c-$th class labelled data samples owned locally by the $k-$th party. Let $\mathcal{P}_c^k$ be the wide conditionally deep membership-mapping autoencoder learned from $\mathbf{Y}^k_c$ and $\mathbf{A}_{\mathcal{P}_c^k}$ be the corresponding fuzzy attribute. The fuzzy classifier~(\ref{eq_738594.721927}) can be extended for distributed setting as follows:
  \begin{IEEEeqnarray}{CCC}
 \nonumber \mbox{If $y$ is $\mathbf{A}_{\mathcal{P}_1^1}$ OR $\mathbf{A}_{\mathcal{P}_1^2}$ OR $\cdots$ OR $\mathbf{A}_{\mathcal{P}_1^K}$} &, & \mbox{then the class is 1};\\
 \label{eq_738595.419} & \vdots & \\
\nonumber  \mbox{If $y$ is $\mathbf{A}_{\mathcal{P}_C^1}$ OR $\mathbf{A}_{\mathcal{P}_C^2}$ OR $\cdots$ OR $\mathbf{A}_{\mathcal{P}_C^K}$} &, & \mbox{then the class is C}.
   \end{IEEEeqnarray}                 
The class-label associated to a data point $y$ is predicted based on fuzzy rules~(\ref{eq_738595.419}) as
  \begin{IEEEeqnarray}{rCl}
\label{eq_738595.4358}\hat{c} & = & \arg \; \max_{1 \leq c \leq C} \; \left( \max_{1 \leq k \leq K} \mu_{ \mathbf{A}_{\mathcal{P}_c^k}}(y) \right) \\ 
\label{eq_738595.726728} & = & \arg \; \min_{1 \leq c \leq C} \; \left( \min_{1 \leq k \leq K} \left( 1 - \mu_{ \mathbf{A}_{\mathcal{P}_c^k}}(y) \right)\right) \\
\label{eq_738597.697428}& = & \arg \; \min_{1 \leq c \leq C} \; \left( \min_{1 \leq k \leq K}  \bar{\mu}_{ \mathbf{A}_{\mathcal{P}_c^k}}(y) \right),
   \end{IEEEeqnarray}      
where $\bar{\mu}_{ \mathbf{A}_{\mathcal{P}_c^k}}(y) = 1 - \mu_{ \mathbf{A}_{\mathcal{P}_c^k}}(y)$. 
\subsection{Secure Homomorphic Evaluation of Global Classifier}
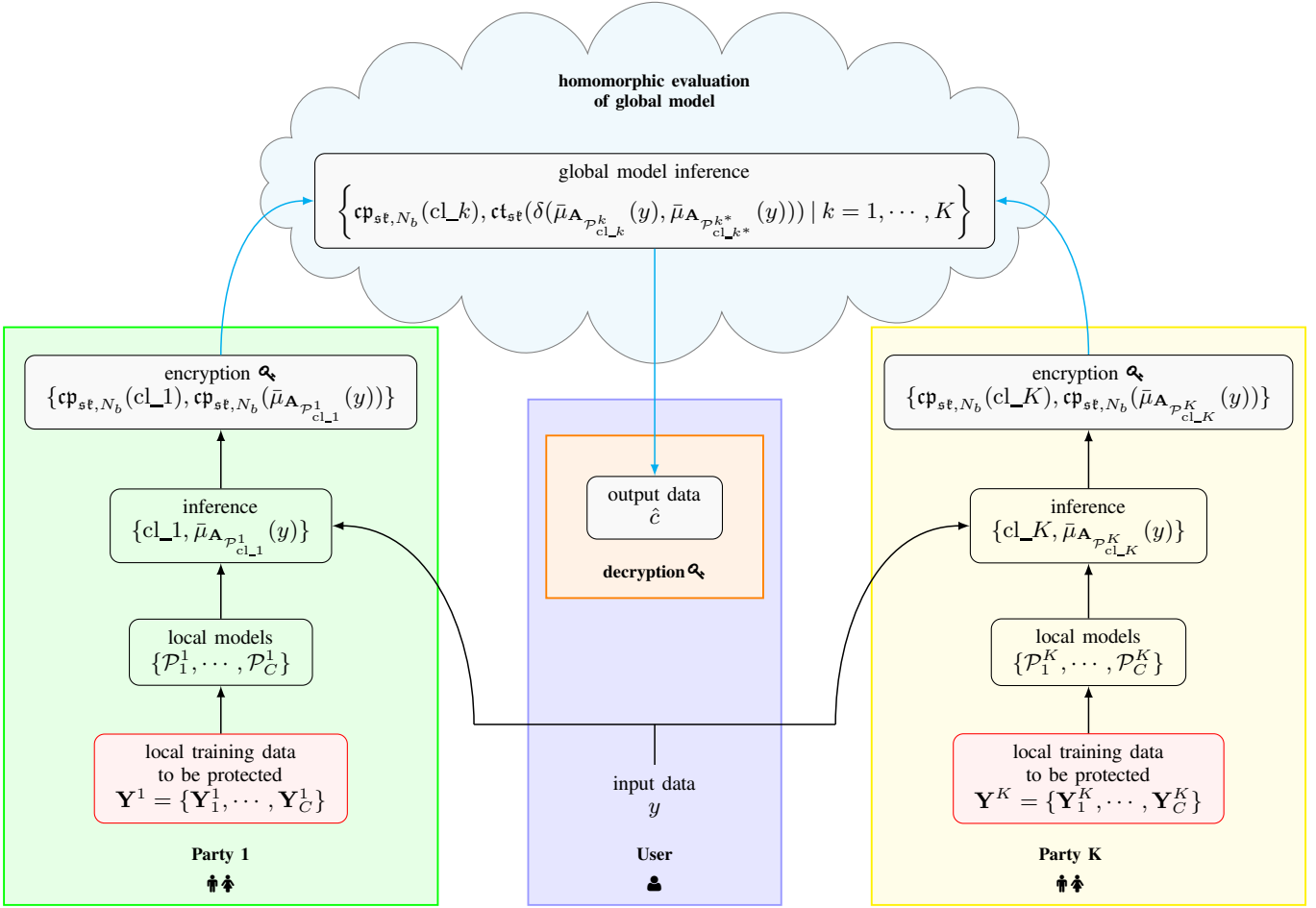
\begin{figure*}
\centering
\begin{tikzpicture}[scale=1]
\path[fill=green!10](-3,-5)--(3,-5)--(3,3)--(-3,3)--cycle;
\draw[green,line width = 0.25mm](-3,-5)--(3,-5)--(3,3)--(-3,3)--cycle;
\draw (0,-4.5) node[]{\bfseries $\begin{array}{c} \mbox{\scriptsize Party 1} \\ \mbox{\scriptsize \faMale\:\faFemale} \end{array}$};
\draw (0,-3.25) node[rounded corners,draw=red,fill=red!5](n1){\footnotesize $\begin{array}{c}\mbox{local training data} \\ \mbox{to be protected} \\ \mbox{\small $\mathbf{Y}^1  =  \{ \mathbf{Y}^1_1,\cdots, \mathbf{Y}^1_C \}$} \end{array}$};
\draw (0,-1.5) node[rounded corners,draw](n2){ \footnotesize $\begin{array}{c}\mbox{local models} \\ \mbox{\small  $ \{ \mathcal{P}_1^1,\cdots, \mathcal{P}_C^1 \}$} \end{array}$};
\draw[-latex,line width=0.2mm] (n1) to [out=90,in=-90] (n2);  
\draw (0,0.25) node[rounded corners,draw](n4){ \footnotesize $\begin{array}{c}\mbox{inference} \\ \mbox{\small $\{\mathrm{cl}\_1,\bar{\mu}_{\mathbf{A}_{ \mathcal{P}_{\mathrm{cl}\_1}^1}}(y)\}$} \end{array}$};
\draw[-latex,line width=0.2mm] (n2) to [out=90,in=-90] (n4);   
\path[fill=yellow!10](9,-5)--(15,-5)--(15,3)--(9,3)--cycle;
\draw[yellow,line width = 0.25mm](9,-5)--(15,-5)--(15,3)--(9,3)--cycle;
\draw (11.75,-4.5) node[]{\bfseries $\begin{array}{c} \mbox{\scriptsize Party K} \\ \mbox{\scriptsize \faMale\:\faFemale} \end{array}$};
\draw (12,0.25) node[rounded corners,draw](n11){ \footnotesize $\begin{array}{c}\mbox{inference} \\ \mbox{\small $\{\mathrm{cl}\_K,\bar{\mu}_{\mathbf{A}_{ \mathcal{P}_{ \mathrm{cl}\_K}^K}}(y)\}$} \end{array}$};
 \draw (12,-1.5) node[rounded corners,draw](n14){ \footnotesize $\begin{array}{c}\mbox{local models} \\ \mbox{\small $ \{ \mathcal{P}_1^K,\cdots, \mathcal{P}_C^K \}$} \end{array}$};
  \draw[-latex,line width=0.2mm] (n14) to [out=90,in=-90] (n11); 
\draw (12,-3.25) node[rounded corners,draw=red,fill= red!5](n15){\footnotesize $\begin{array}{c}\mbox{local training data} \\ \mbox{to be protected} \\ \mbox{\small $\mathbf{Y}^K  =  \{ \mathbf{Y}^K_1,\cdots, \mathbf{Y}^K_C \}$} \end{array}$};
\draw[-latex,line width=0.2mm] (n15) to [out=90,in=-90] (n14);  
\path[fill=blue!10](4.25,-5)--(7.75,-5)--(7.75,2)--(4.25,2)--cycle;   
\draw[blue!40,line width = 0.25mm](4.25,-5)--(7.75,-5)--(7.75,2)--(4.25,2)--cycle; 
\draw (6,-4.5) node[]{\bfseries $\begin{array}{c} \mbox{\scriptsize User} \\ \mbox{\scriptsize \faUser} \end{array}$};
  \draw (6,-3.5) node[](n20){ \footnotesize $\begin{array}{c}\mbox{input data} \\ \mbox{\small $y$} \end{array}$};
\draw[thick,line width=0.2mm](n20) -- (6,-2.5);
\draw[thick,line width=0.2mm](3.5,-2.5) -- (8.5,-2.5);
\draw[-latex,thick,line width=0.2mm] (3.5,-2.5) to [out=90,in=0] (n4);   
 \draw[-latex,thick,line width=0.2mm] (8.5,-2.5) to [out=90,in=180] (n11); 
\draw (0,2.1) node[rounded corners,draw, fill = gray!5](n7){ \footnotesize $\begin{array}{c}\mbox{encryption {\scriptsize \faKey}} \\ \mbox{\small $\{\mathfrak{cp}_{\mathfrak{sk},N_b}(\mathrm{cl}\_1), \mathfrak{cp}_{\mathfrak{sk},N_b}(\bar{\mu}_{\mathbf{A}_{ \mathcal{P}_{\mathrm{cl}\_1}^1}}(y))\}$} \end{array}$};
  \draw[-latex,line width=0.2mm] (n4) to [out=90,in=-90] (n7); 
\draw (12,2.1) node[rounded corners,draw, fill = gray!5](n9){ \footnotesize $\begin{array}{c}\mbox{encryption {\scriptsize \faKey}} \\ \mbox{\small $\{\mathfrak{cp}_{\mathfrak{sk},N_b}(\mathrm{cl}\_K), \mathfrak{cp}_{\mathfrak{sk},N_b}(\bar{\mu}_{\mathbf{A}_{ \mathcal{P}_{\mathrm{cl}\_K}^K}}(y))\}$} \end{array}$};
  \draw[-latex,line width=0.2mm] (n11) to [out=90,in=-90] (n9); 
  \node[cloud,
    draw = gray,
    fill = cyan!5,
    minimum width = 11cm,
    minimum height = 5cm,
    cloud puffs = 18] (c) at (6,5) {};
\draw (6,4.75) node[rounded corners,draw, fill=gray!5](n8){ \footnotesize $\begin{array}{c}\mbox{global model inference} \\ \mbox{\small $\left\{\mathfrak{cp}_{\mathfrak{sk},N_b}(\mathrm{cl}\_k), \mathfrak{ct}_{\mathfrak{sk}}(\delta(\bar{\mu}_{\mathbf{A}_{ \mathcal{P}_{\mathrm{cl}\_k}^k}}(y), \bar{\mu}_{\mathbf{A}_{ \mathcal{P}_{\mathrm{cl}\_{k^*}}^{k^*}}}(y) ))\: | \: k = 1,\cdots,K\right \}$} \end{array}$};
\draw[-latex,line width=0.2mm,cyan] (n7) to [out=90,in=180] (n8);  
\draw (6,6.25) node[]{\bfseries {\scriptsize $\begin{array}{c} \mbox{homomorphic evaluation} \\ \mbox{of global model} \end{array}$  }};
\draw[-latex,line width=0.2mm,cyan] (n9) to [out=90,in=0] (n8);  
\path[fill=orange!10](4.5,-0.75)--(7.5,-0.75)--(7.5,1.5)--(4.5,1.5)--cycle;
 \draw[orange,line width = 0.25mm] (4.5,-0.75)--(7.5,-0.75)--(7.5,1.5)--(4.5,1.5)--cycle;
\draw (6,0.5) node[rounded corners,draw, fill = gray!5](n17){ \footnotesize $\begin{array}{c}\mbox{output data} \\ \mbox{\small $\hat{c}$} \end{array}$};
\draw (6,-0.4) node[]{\bfseries {\scriptsize decryption\,}{\scriptsize \faKey}};
\draw[-latex,line width=0.2mm,cyan] (n8) to [out=-90,in=90] (n17);  
\end{tikzpicture}
\caption{A practical method for secure distributed deep learning based on membership-mappings and fully homomorphic encryption.}\label{figure_secure_distributed_deep_learning}
\end{figure*} 
For a secure distributed learning based on fully homomorphic encryption, a practical approach is suggested based on the observation that instead of encrypting higher-dimensional vectors, it is sufficient to encrypt the scalar values $\{\bar{\mu}_{ \mathbf{A}_{\mathcal{P}_c^k}}(y) \; | \; c = 1,\cdots,C; \: k = 1,\cdots,K\}$ for the homomorphic evaluation of classifier~(\ref{eq_738595.419}) (i.e. evaluation of~(\ref{eq_738597.697428})) in an efficient manner. That is, the homomorphic evaluation of the ``$\arg \; \min$'' function over $K \times C$ encrypted values is required. More efficiently, the homomorphic evaluation of the ``$\arg \; \min$'' function over $K$ encrypted values is sufficient for the homomorphic evaluation of the classifier. For this, define $\mathrm{cl}\_k$ as the class-label predicted by the $k-$th local classifier, i.e., 
  \begin{IEEEeqnarray}{rCl}
\mathrm{cl}\_k & = &    \mathcal{C}(y;\{ \mathcal{P}_c^{k}\}_{c=1}^C) 
   \end{IEEEeqnarray}    
where $ \mathcal{C}(\cdot)$ is defined as in (\ref{eq_738594.728862}). Now, (\ref{eq_738597.697428}) can be alternatively expressed as
  \begin{IEEEeqnarray}{rCl}
 \label{eq_738598.8647} \hat{c} & = & \mathrm{cl}\_{k^*}  \\
\label{eq_738596.584220}  k^* & = & \arg \; \min_{1 \leq k \leq K}\;   \bar{\mu}_{\mathbf{A}_{ \mathcal{P}_{ \mathrm{cl}\_k}^k}}(y). 
   \end{IEEEeqnarray} 
Let $\delta(\mathfrak{m}_1,\mathfrak{m}_2)$ be the Kronecker delta function of two variables $\mathfrak{m}_1,\mathfrak{m}_2 \in [0,1]$ defined as
\begin{IEEEeqnarray}{rCl}
\delta(\mathfrak{m}_1,\mathfrak{m}_2) & = & \left\{ \,
    \begin{IEEEeqnarraybox}[][c]{l?s}
      \IEEEstrut
     1 & if $\mathfrak{m}_1=\mathfrak{m}_2$, \\
     0 & if $\mathfrak{m}_1 \neq \mathfrak{m}_2$.
      \IEEEstrut
    \end{IEEEeqnarraybox}
\right.
\end{IEEEeqnarray}    
It follows from (\ref{eq_738598.8647}-\ref{eq_738596.584220}) that 
\begin{IEEEeqnarray}{rCl}
\label{eq_738598.8701} \hat{c} & = & \sum_{k=1}^K \mathrm{cl}\_k \; \delta \left(\bar{\mu}_{\mathbf{A}_{ \mathcal{P}_{ \mathrm{cl}\_k}^k}}(y),\bar{\mu}_{\mathbf{A}_{ \mathcal{P}_{ \mathrm{cl}\_{k^*}}^{k^*}}}(y)\right).
\end{IEEEeqnarray}
 A practical method using TFHE scheme~\cite{10.1007/978-3-662-53887-6_1,10.1007/978-3-319-70694-8_14} is provided for secure distributed deep learning. For this, define the followings: 
\begin{itemize}
\item For a given positive integer $N_b \in \mathbb{Z}_{>0}$, let $\mathfrak{pt}_{N_b}:[0,1]\rightarrow \{0,1,\cdots,2^{N_b}-1 \}$ be a function defined as
\begin{IEEEeqnarray}{rCl}
 \mathfrak{pt}_{N_b}(\mathfrak{m}) & := & \lceil (2^{N_b}-1) \mathfrak{m} \rceil,\;  \mathfrak{m} \in [0,1].
\end{IEEEeqnarray}     
In our setting, $\mathfrak{pt}_{N_b}( \mathfrak{m})$ is the plaintext that encodes a message $ \mathfrak{m} \in [0,1]$ as unsigned $N_b-$bit integer. 
\item Let $\mathrm{BitDec}_{N_b}: \{0,1,\cdots,2^{N_b}-1 \} \rightarrow \{0,1\}^{N_b}$ be the binary representation of a $N_b-$bit unsigned integer. That is,
\begin{IEEEeqnarray}{rCl}
\left(\mathfrak{bt}_1( \mathfrak{m}),\cdots,\mathfrak{bt}_{N_b}( \mathfrak{m})\right) &  = & \mathrm{BitDec}_{N_b}(\mathfrak{pt}_{N_b}( \mathfrak{m})), \IEEEeqnarraynumspace
\end{IEEEeqnarray}  
where $\mathfrak{bt}_k( \mathfrak{m}) \in \{0,1\}$ for all $k \in \{1,2,\cdots,N_b\}$.  
\item Let $N_c$ be the ciphertext dimension set for a given value of security bits, say 128 bits security.  
\item Let $\mathfrak{sk} \in \{0,1\}^{N_c}$ be a secret key generated for TFHE encryption.
\item  Let $\mathfrak{ct}_{\mathfrak{sk}}(\mathfrak{bt}) \in  \mathbb{T}^{N_c+1}$, where $\mathbb{T} = \mathbb{R}/\mathbb{Z}$, be the TFHE encryption of a bit $\mathfrak{bt} \in \{0,1 \}$, i.e., $\mathfrak{ct}_{\mathfrak{sk}}(\mathfrak{bt}) = \mathrm{TFHE.Enc}(\mathfrak{bt};\mathfrak{sk})$.
\item Let $\mathfrak{cp}_{\mathfrak{sk},N_b}:[0,1] \rightarrow \mathbb{T}^{N_b (N_c+1)}$ be a function defined as
\begin{IEEEeqnarray}{rCl}
\mathfrak{cp}_{\mathfrak{sk},N_b}(\mathfrak{m}) & := & \left(\mathfrak{ct}_{\mathfrak{sk}}\left(\mathfrak{bt}_1( \mathfrak{m})\right),\cdots,\mathfrak{ct}_{\mathfrak{sk}}\left(\mathfrak{bt}_{N_b}( \mathfrak{m})\right)\right) \IEEEeqnarraynumspace
\end{IEEEeqnarray} 
where $\mathfrak{ct}_{\mathfrak{sk}}\left(\mathfrak{bt}_k( \mathfrak{m})\right)$ is the TFHE encryption of bit $\mathfrak{bt}_k( \mathfrak{m})$. Thus, $\mathfrak{cp}_{\mathfrak{sk},N_b}(\mathfrak{m})$ homomorphically encrypts the message $\mathfrak{m} \in [0,1]$ with $N_b$-bit precision.  
\end{itemize}       
Our approach to homomorphically evaluate the global classifier~(\ref{eq_738595.419}) is shown in Fig.~\ref{figure_secure_distributed_deep_learning}. The approach consists of following steps:
\begin{enumerate}
\item A pair of secret and cloud keys is generated. The secret key is meant for encryption and decryption. The cloud key is exported to the cloud, and allows to operate over encrypted data.
\item For a given input $y$, the outputs of local classifiers $\{\mathfrak{cp}_{\mathfrak{sk},N_b}(\mathrm{cl}\_k), \mathfrak{cp}_{\mathfrak{sk},N_b}(\bar{\mu}_{\mathbf{A}_{ \mathcal{P}_{\mathrm{cl}\_k}^k}}(y))\; | \; k = 1,\cdots,K\}$ are sent to the cloud for performing secure homomorphic computations.
 \item The values $\{  \delta (\bar{\mu}_{\mathbf{A}_{ \mathcal{P}_{ \mathrm{cl}\_k}^k}}(y),\bar{\mu}_{\mathbf{A}_{ \mathcal{P}_{ \mathrm{cl}\_{k^*}}^{k^*}}}(y))  \; |  \: k = 1,\cdots,K\}$ are homomorphically evaluated in the cloud from the encrypted data (sent by different parties) using a boolean circuit composed of bootstrapped binary gates.
 \item The encrypted output of the global model $\{\mathfrak{cp}_{\mathfrak{sk},N_b}(\mathrm{cl}\_k), \mathfrak{ct}_{\mathfrak{sk}}(\delta(\bar{\mu}_{\mathbf{A}_{ \mathcal{P}_{\mathrm{cl}\_k}^k}}(y), \bar{\mu}_{\mathbf{A}_{ \mathcal{P}_{\mathrm{cl}\_{k^*}}^{k^*}}}(y) ))\: | \: k = 1,\cdots,K\}$ is sent to the user.  
\item The class-label associated to a data point $y$ is predicted using (\ref{eq_738598.8701}) after decrypting the data provided by the cloud.  
\end{enumerate}
\section{Experiments}\label{sec_experiments}
The proposed method was implemented using MATLAB R2017b and TFHE C/C++ library~\cite{TFHE} on a MacBook Pro machine with a 2.2 GHz Intel Core i7 processor and 16 GB of memory. The previous works \cite{8888203,10.1007/978-3-030-87101-7_14,KUMAR20211} have already verified the competitive performance of membership-mappings in classification applications. In this study, our focus is to verify the application potential of the proposed approach and to compare the method with the alternative differentially private distributed deep learning approach~\cite{KUMAR202187,10.1145/3386392.3399562}. 
\subsection{Details}
Under a distributed deep learning scenario, the local models are developed using Algorithm~\ref{algorithm_optimal_RWDSFMA} taking $L = 5$, $n = 20$, and possible $r_{max}$ values array as $\{0.5\}$. Targeting 128-bits of security, TFHE library is used to homomorphically evaluate the global classifier with the precision of 16-bits and also 8-bits. For a comparison, differentially private local models, developed using Algorithm~\ref{algorithm_optimal_RWDSFMA} on the noisy training data obtained via optimal noise adding mechanism~\cite{KUMAR202187,10.1145/3386392.3399562} (taking adjacency parameter $d = 1$, failure probability $\delta = 1\mathrm{e}{-5}$, and privacy-loss bound $\epsilon = 0.1$ and also $\epsilon = 1$), are combined through fuzzy rules (\ref{eq_738595.419}) without any encryption. The test data accuracy and average computational time required for secure homomorphic computations in the cloud (for computing the encrypted global output for a given input) are considered as performance indices.          
\subsection{MNIST Dataset}
The first experiment is on the widely used MNIST digits dataset containing $28 \times 28$ sized images divided into training set of 60000 images and testing set of 10000 images. The images' pixel values were divided by 255 to normalize the values in the range from $0$ to $1$. The $28 \times 28$ normalized values of each image were flattened to an equivalent $784-$dimensional data vector. A two-party scenario is considered such that Party-A owns all the training images of odd digits while Party-B owns the rest training images of even digits.          
\begin{table}
\caption{Experimental Results on MNIST Dataset\label{table_mnist_results}}
\centering
\begin{tabular}{|c||c||c|}
\hline
\bfseries Method & \bfseries Accuracy & \bfseries Time  \\
\hline
$\begin{array}{c} \mbox{Proposed (8-bits precision of} \\ \mbox{homomorphic computations)} \end{array}$ & 0.9762 & 3.2251 s\\
\hline
$\begin{array}{c} \mbox{Proposed (16-bits precision of} \\ \mbox{homomorphic computations)} \end{array}$ & 0.9872 & 4.9785 s \\
\hline
$\begin{array}{c} \mbox{differentially private} \\ \mbox{distributed deep learning with} \\ \mbox{privacy-loss bound $\epsilon = 0.1$ \cite{KUMAR202187}} \end{array}$ & 0.0892 & n/a \\
\hline
$\begin{array}{c} \mbox{differentially private} \\ \mbox{distributed deep learning with} \\ \mbox{privacy-loss bound $\epsilon = 1$ \cite{KUMAR202187}} \end{array} $ & 0.8994 & n/a \\
\hline
\end{tabular}
\end{table}
Table~\ref{table_mnist_results} reports the experimental results.  
\subsection{Freiburg Groceries Dataset}
The second experiment is on ``Freiburg Groceries Dataset'' considered previously~\cite{KUMAR202187} for privacy-preserving distributed learning experiments. The dataset contains 4947 labeled images of grocery products categorized into 25 different classes. A feature vector is created from each image by extracting features from ``AlexNet'' and ``VGG-16'' networks which are pre-trained Convolutional Neural Networks. The activations of the fully connected layer ``fc6'' in AlexNet constitute a $4096-$dimensional feature vector. Similarly, the activations of the fully connected layer ``fc6'' in VGG-16 constitute another $4096-$dimensional feature vector. The features extracted by both networks are joined together to form a $8192-$dimensional vector. The feature vectors are normalized to have zero-mean and unity-variance along each dimension. The set of normalized feature vectors is split into a training set containing around 80\% of data points and a testing set containing remaining data points. A three-party scenario is created such that Party-A owns all the training data of first ten grocery categories, Party-B owns all the training data of second ten grocery categories, and Party-C owns all the training data of rest five grocery categories.   
\begin{table}
\caption{Experimental Results on Freiburg Groceries Dataset\label{table_groceries_results}}
\centering
\begin{tabular}{|c||c||c|}
\hline
\bfseries Method & \bfseries Accuracy & \bfseries Time  \\
\hline
$\begin{array}{c} \mbox{Proposed (8-bits precision of} \\ \mbox{homomorphic computations)} \end{array}$ & 0.8861 & 4.9534 s\\
\hline
$\begin{array}{c} \mbox{Proposed (16-bits precision of} \\ \mbox{homomorphic computations)} \end{array}$ & 0.8880  & 7.9240 s \\
\hline
$\begin{array}{c} \mbox{differentially private} \\ \mbox{distributed deep learning with} \\ \mbox{privacy-loss bound $\epsilon = 0.1$ \cite{KUMAR202187}} \end{array}$ & 0.1356 & n/a \\
\hline
$\begin{array}{c} \mbox{differentially private} \\ \mbox{distributed deep learning with} \\ \mbox{privacy-loss bound $\epsilon = 1$ \cite{KUMAR202187}} \end{array}$ & 0.8261 & n/a \\
\hline
\end{tabular}
\end{table}
Table~\ref{table_groceries_results} reports the experimental results.  
\subsection{A Biomedical Application}
As an application example, the mental stress detection problem is considered. The dataset from~\cite{9216097}, consisting of heart rate interval measurements of different subjects, is considered for the study of individual stress detection problem. The problem is concerned with the detection of stress on an individual based on the analysis of recorded sequence of R-R intervals, $\{RR^i\}_i$. The R-R data vector at $i-$th time-index, $y^i$, is defined as $y^i  =  \left[\begin{IEEEeqnarraybox*}[][c]{,c/c/c/c,} RR^i & RR^{i-1} & \cdots & RR^{i-d}\end{IEEEeqnarraybox*} \right]^T$. That is, the current interval and history of previous $d$ intervals constitute the data vector. Assuming an average heartbeat of 72 beats per minute, $d$ is chosen as equal to $72 \times 3 = 216$ so that R-R data vector consists of on an average 3-minutes long R-R intervals sequence. A dataset, say $\{y^i\}_{i}$, is built via 1) preprocessing the R-R interval sequence $\{RR^i\}_i$ with an impulse rejection filter for artifacts detection, and 2) excluding the R-R data vectors containing artifacts from the dataset. The dataset contains the stress-score on a scale from 0 to 100. A label of either ``\emph{no-stress}'' or ``\emph{under-stress}'' is assigned to each $y^i$ based on the stress-score. Thus, we have a binary classification problem. A two-party collaborative learning scenario is considered where a randomly chosen subject is considered as Party-A. While keeping Party-A fixed, the distributed learning experiments are performed independently on every other subject being considered as Party-B. For each subject, 50\% of the data samples serve as training data while remaining as test data. The subjects, with data containing both the classes and at least 60 samples, are considered for experimentation. There are in total 48 such subjects.  
\begin{table}
\caption{Experimental Results on Heart Rate Variability Dataset\label{table_hrv_results}}
\centering
\begin{tabular}{|c||c||c|}
\hline
\bfseries Method & \bfseries Accuracy & \bfseries Time  \\
\hline
$\begin{array}{c} \mbox{Proposed (8-bits precision of} \\ \mbox{homomorphic computations)} \end{array}$ & 0.8358  & 3.1571 s\\
\hline
$\begin{array}{c} \mbox{Proposed (16-bits precision of} \\ \mbox{homomorphic computations)} \end{array}$ &  0.9580 & 4.8610 s \\
\hline
$\begin{array}{c} \mbox{differentially private} \\ \mbox{distributed deep learning with} \\ \mbox{privacy-loss bound $\epsilon = 0.1$ \cite{KUMAR202187}} \end{array}$ &  0.5123 & n/a \\
\hline
$\begin{array}{c} \mbox{differentially private} \\ \mbox{distributed deep learning with} \\ \mbox{privacy-loss bound $\epsilon = 1$ \cite{KUMAR202187}} \end{array}$ & 0.6873 & n/a \\
\hline
\end{tabular}
\end{table}
The experimental results, averaged over 48 independent experiments, are reported in Table~\ref{table_hrv_results}.
\subsection{Scalability}
The computational time required for the homomorphic evaluation of global model depends on the number of parties (i.e. $K$) participating in collaborative learning. Therefore, experiments are performed to study the computational time as the number of parties is varied from $K = 2$ to $K = 100$.         
\begin{figure}
\centering
\includegraphics[width=0.5\textwidth]{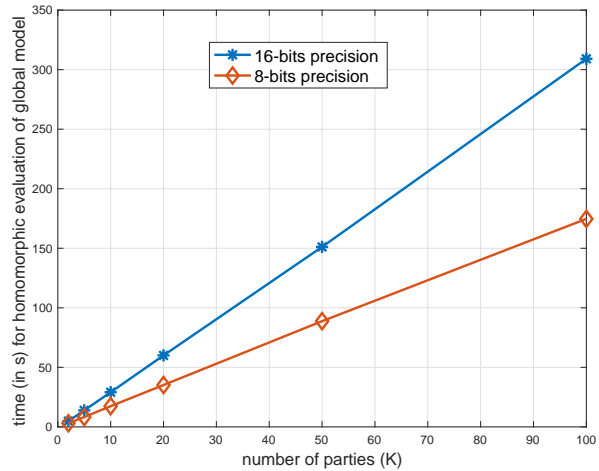}
\caption{Computational time vs. number of parties.}\label{fig_computational_time}
\end{figure}  
Fig.~\ref{fig_computational_time} plots the computational time required for secure homomorphic computations on a MacBook Pro machine with a 2.2 GHz Intel Core i7 processor and 16 GB of memory without using parallel computing. The ratio of computational time to the number of parties is observed to be approximately equal to 3 s for 16-bit precision and 1.75 s for 8-bit precision.    
\subsection{Evaluation of the Proposed Approach}
To evaluate the gains achieved in accuracy and computational time as a result of the proposed approach to secure distributed deep learning (summarized in Fig.~\ref{figure_secure_distributed_deep_learning}), a comparison is made with a variant of the TFHE fully homomorphic encryption scheme~\cite{DBLP:conf/cscml/ChillottiJP21}. The study in \cite{DBLP:conf/cscml/ChillottiJP21} reports the experiments on MNIST dataset for evaluating the neural networks with different depths (referred to as NN-20, NN-50, and NN-100) over TFHE fully homomorphically encrypted data. The results of \cite{DBLP:conf/cscml/ChillottiJP21} are compared with the proposed method (considering two-party scenario with Party-A owning odd-digit images and Party-B owning even-digit images) in Table~\ref{table_comparison}.     
\begin{table}
\caption{Gains achieved by the proposed approach on MNIST dataset\label{table_comparison}}
\centering     
\begin{tabular}{|c||c||c||c||c|}
\hline
\bfseries Method & \bfseries Security & \bfseries Machine & \bfseries Accuracy & \bfseries Time \\
\hline
Proposed & 128-bits & $\begin{array}{c} \mbox{2.2 GHz} \\ \mbox{Intel Core i7} \end{array}$ & 0.987 & 4.98 s \\
\hline
NN-20~\cite{DBLP:conf/cscml/ChillottiJP21} & 128-bits & $\begin{array}{c} \mbox{2.6 GHz} \\ \mbox{Intel Core i7} \end{array}$ & 0.971 &   115.52 s \\ 
\hline   
NN-50~\cite{DBLP:conf/cscml/ChillottiJP21} & 128-bits & $\begin{array}{c} \mbox{2.6 GHz} \\ \mbox{Intel Core i7} \end{array}$ & 0.947 &   233.55 s \\ 
\hline  
NN-100~\cite{DBLP:conf/cscml/ChillottiJP21} & 128-bits & $\begin{array}{c} \mbox{2.6 GHz} \\ \mbox{Intel Core i7} \end{array}$ & 0.830 &   481.61 s \\ 
\hline  
\end{tabular}
\end{table}    
\subsection{Main Results}
Following inferences are made from the results of aforementioned experiments.
\begin{itemize}
\item As expected, the proposed approach leads to better accuracy (observed in Tables \ref{table_mnist_results}, \ref{table_groceries_results}, \ref{table_hrv_results}) than the differentially private approach~\cite{KUMAR202187}, since differential privacy requires contaminating data with noise to preserve privacy.  
\item The proposed membership-mappings based approach is capable of handling the large computational overhead issue of fully homomorphic encryption, since the computational time on a MacBook Pro machine with a 2.2 GHz Intel Core i7 processor and 16 GB of memory (as reported in Tables \ref{table_mnist_results}, \ref{table_groceries_results}, \ref{table_hrv_results}, \ref{table_comparison}) is practical.      
\item A linear increase of the computational time with increasing number of parties, as observed in Fig.~\ref{fig_computational_time}, indicates that the proposed approach is scalable using parallel computing.   
\item Remarkably, the computational time required for secure homomorphic evaluation of the global model in the cloud is independent of the dimension of the input data, and thus the approach is practical. The computational time depends only on the number of parties and the chosen precision. The ratio of computational time to the number of parties as approximately equal to 3 s for 16-bit precision verifies the application potential. 
\end{itemize}

\section{Concluding Remarks}\label{sec_conclusion}
This study has outlined a membership-mappings based approach to secure distributed deep learning. The crux of our methodology lies in defining fuzzy attributes (which are induced by globally convergent and robust variational membership-mappings based local deep models) allowing to combine local models by means of a rule-based fuzzy system, thus facilitating the homomorphic evaluation of the global model efficiently. The feature, that the computational time for secure homomorphic evaluation of the global model in the cloud is independent of the dimension of input data, adds to the practicality of the approach. The experimental results verify that the proposed method, while preserving the privacy in a distributed learning scenario using fully homomorphic encryption, remains accurate, practical, and scalable.

{\appendices
\section{Membership-Mapping Output Estimation}\label{appendix_738242.626259}
Using (\ref{eq_pf_1001_student_t}) and (\ref{eq_pf_1002}), we have 
 \begin{IEEEeqnarray}{rCl}
\label{eq_satguru_3}\left< (\mathrm{f}_j)_i\right>_{\mu_{\mathrm{f}_j;\mathrm{u}_j } } & = & (K_{\mathrm{x}\mathrm{a}} K_{\mathrm{a}\mathrm{a}}^{-1}  \mathrm{u}_j )_i \\
\label{eq_satguru_4}& = & G(x^i) K_{\mathrm{a}\mathrm{a}}^{-1}  \mathrm{u}_j.
 \end{IEEEeqnarray}
Thus,   
 \begin{IEEEeqnarray}{rCl}
\label{eq_satguru_5} & \widehat{ \mathcal{F}_j(x^{i}) } = &  G(x^i)K_{\mathrm{a}\mathrm{a}}^{-1} \left< \mathrm{u}_j \right>_{\mu^*_{\mathrm{u}_j}}.
 \end{IEEEeqnarray}  
Using (\ref{eq_q_u_vec_optimal}) and (\ref{eq_hat_m_u_1000}) in (\ref{eq_satguru_5}), we have
\begin{IEEEeqnarray}{rCl}
\label{eq_satguru_11}  \widehat{ \mathcal{F}_j(x^{i})}   &=&\beta  \left(G(x^i) \right) K_{\mathrm{a}\mathrm{a}}^{-1} \hat{K}_{\mathrm{u}_j} K_{\mathrm{a}\mathrm{a}}^{-1}K_{\mathrm{x}\mathrm{a}}^T \mathrm{y}_j .
 \end{IEEEeqnarray} 
Substituting $\hat{K}_{\mathrm{u}_j}$ from (\ref{eq_hat_K_u_1000}) in (\ref{eq_satguru_11}), we get (\ref{eq_satguru_12}). 

\section{Proof of Result~\ref{result_definition_variance_function}}\label{appendix_738612.5557}
The proof is split into three parts.
\paragraph{Part 1}
Since $K_{\mathrm{a}\mathrm{a}} > 0$, there exists the unique square root, $K_{\mathrm{a}\mathrm{a}}^{1/2} > 0$. Thus,
\begin{IEEEeqnarray}{rCl}
S V^T  K_{\mathrm{a}\mathrm{a}}^{-1} VS & = & \left(K_{\mathrm{a}\mathrm{a}}^{-1/2}VS \right)^T\left( K_{\mathrm{a}\mathrm{a}}^{-1/2}VS \right) \\
\label{eq_738569.4944}& > & 0.
\end{IEEEeqnarray} 
Since $\tau > 0$ and $\beta > 0$,
\begin{IEEEeqnarray}{rCl}
\text{min\_eigen}\left(I + \frac{1}{(\tau + \beta^{-1})} S V^T  K_{\mathrm{a}\mathrm{a}}^{-1} VS \right) & > & 1 
\end{IEEEeqnarray} 
where ``$\text{min\_eigen}(\cdot)$'' denotes the minimum eigenvalue. Thus,
\begin{IEEEeqnarray}{rCl}
\label{eq_738574.58}\text{max\_eigen}\left(\left(I + \frac{1}{(\tau + \beta^{-1})} S V^T  K_{\mathrm{a}\mathrm{a}}^{-1} VS \right)^{-2} \right) & < & 1 \IEEEeqnarraynumspace 
\end{IEEEeqnarray} 
where ``$\text{max\_eigen}(\cdot)$'' denotes the maximum eigenvalue. As a result of (\ref{eq_738574.58}), 
\begin{IEEEeqnarray}{rCl}
\label{eq_738575.4658} \mathcal{R}(\beta^{-1}) &  < &  \frac{1}{pN}\sum_{j=1}^p(\|  b_j^1 \|^2 +\|  b_j^2 \|^2).
\end{IEEEeqnarray}
As $U$ is orthogonal, it follows from (\ref{eq_738574.5863}) that $ \|  b_j^1 \|^2 +\|  b_j^2 \|^2 = \| \mathrm{y}_j  \|^2$, and thus
\begin{IEEEeqnarray}{rCl}
\mathcal{R}(\beta^{-1}) &  < &  \beta^{-1}|_{up}.
\end{IEEEeqnarray}
It follows immediately from (\ref{eq_738570.8644}) that $\mathcal{R}(\beta^{-1})   >   \beta^{-1}|_{low}$. Hence, $\mathcal{R}(\beta^{-1}) \in \left(\beta^{-1}|_{low}, \beta^{-1}|_{up}  \right)$. 
\paragraph{Part 2}
The derivative of $\mathcal{R}$ w.r.t. $\beta^{-1}$ is given as
\begin{IEEEeqnarray}{rCl}
\label{eq_738569.4536} \lefteqn{\frac{\dd \mathcal{R}(\beta^{-1})}{\dd \beta^{-1}}  = }\\
 \nonumber && \frac{2}{pN}\sum_{j=1}^p \left \{ (\tau + \beta^{-1})(b_j^1)^T \left( (\tau + \beta^{-1})I + S V^T  K_{\mathrm{a}\mathrm{a}}^{-1} VS   \right)^{-2}  b_j^1 \right. \\
 \nonumber && \left.  {-} \:   (\tau + \beta^{-1})^2 (b_j^1)^T \left( (\tau + \beta^{-1})I + S V^T  K_{\mathrm{a}\mathrm{a}}^{-1} VS   \right)^{-3}  b_j^1 \right \}.
  \end{IEEEeqnarray}
Consider
\begin{IEEEeqnarray}{rCl}
\nonumber \lefteqn{(\tau + \beta^{-1})^2 (b_j^1)^T \left( (\tau + \beta^{-1})I + S V^T  K_{\mathrm{a}\mathrm{a}}^{-1} VS   \right)^{-3}  b_j^1} \\
 & \leq & (\tau + \beta^{-1})^2 \left \| \left( (\tau + \beta^{-1})I + S V^T  K_{\mathrm{a}\mathrm{a}}^{-1} VS   \right)^{-1} \right \|_2 \\
 \nonumber && {\times} \: (b_j^1)^T \left( (\tau + \beta^{-1})I + S V^T  K_{\mathrm{a}\mathrm{a}}^{-1} VS   \right)^{-2} b_j^1 \\
 & = & (\tau + \beta^{-1}) \left \| \left( I + \frac{1}{(\tau + \beta^{-1})} S V^T  K_{\mathrm{a}\mathrm{a}}^{-1} VS   \right)^{-1} \right \|_2 \\
 \nonumber && {\times} \: (b_j^1)^T \left( (\tau + \beta^{-1})I + S V^T  K_{\mathrm{a}\mathrm{a}}^{-1} VS   \right)^{-2} b_j^1 \\
\label{eq_738569.4472} & = &  (\tau + \beta^{-1}) \frac{(b_j^1)^T \left( (\tau + \beta^{-1})I + S V^T  K_{\mathrm{a}\mathrm{a}}^{-1} VS   \right)^{-2} b_j^1}{\text{min\_sing}\left( I + \frac{1}{(\tau + \beta^{-1})} S V^T  K_{\mathrm{a}\mathrm{a}}^{-1} VS \right)}\IEEEeqnarraynumspace 
\end{IEEEeqnarray}
where ``$\text{min\_sing}(\cdot)$'' denotes the minimum singular value. Observing that $\tau > 0$, $\beta^{-1} > 0$, and $S V^T  K_{\mathrm{a}\mathrm{a}}^{-1} VS > 0$ (i.e. (\ref{eq_738569.4944})), we have
\begin{IEEEeqnarray}{rCl}
\nonumber \lefteqn{\text{min\_sing}\left( I + \frac{1}{(\tau + \beta^{-1})} S V^T  K_{\mathrm{a}\mathrm{a}}^{-1} VS \right)} \\
\nonumber & = & 1 + \text{min\_sing}\left(\frac{1}{(\tau + \beta^{-1})} S V^T  K_{\mathrm{a}\mathrm{a}}^{-1} VS \right) \\
\label{eq_738569.4493}& > & 1.
\end{IEEEeqnarray}
Combining (\ref{eq_738569.4493}) and (\ref{eq_738569.4472}), we have
\begin{IEEEeqnarray}{rCl}
\label{eq_738569.4533}\lefteqn{(\tau + \beta^{-1})^2 (b_j^1)^T \left( (\tau + \beta^{-1})I + S V^T  K_{\mathrm{a}\mathrm{a}}^{-1} VS   \right)^{-3}  b_j^1}\\
\nonumber & < &  (\tau + \beta^{-1})(b_j^1)^T \left( (\tau + \beta^{-1})I + S V^T  K_{\mathrm{a}\mathrm{a}}^{-1} VS   \right)^{-2} b_j^1. 
\end{IEEEeqnarray}
Using (\ref{eq_738569.4533}) in (\ref{eq_738569.4536}), we get
\begin{IEEEeqnarray}{rCl}
\label{eq_738573.4614} \frac{\dd \mathcal{R}(\beta^{-1})}{\dd \beta^{-1}}  & >  &0.
  \end{IEEEeqnarray}
Observing that $\tau > 0$, $\beta^{-1} > 0$, and $S V^T  K_{\mathrm{a}\mathrm{a}}^{-1} VS > 0$ (i.e. (\ref{eq_738569.4944})), we also have
\begin{IEEEeqnarray}{rCl}
\label{eq_738569.515} (\tau + \beta^{-1})^2 (b_j^1)^T \left( (\tau + \beta^{-1})I + S V^T  K_{\mathrm{a}\mathrm{a}}^{-1} VS   \right)^{-3}  b_j^1 & > & 0. \IEEEeqnarraynumspace
\end{IEEEeqnarray}
Using (\ref{eq_738569.515}) in (\ref{eq_738569.4536}), we get
\begin{IEEEeqnarray}{rCl}
\label{eq_738569.5889} \lefteqn{ \frac{\dd \mathcal{R}(\beta^{-1})}{\dd \beta^{-1}} <}\\
\nonumber  & & \frac{2}{pN} (\tau + \beta^{-1}) \sum_{j=1}^p (b_j^1)^T \left( (\tau + \beta^{-1})I + S V^T  K_{\mathrm{a}\mathrm{a}}^{-1} VS   \right)^{-2}  b_j^1.
  \end{IEEEeqnarray}
Inequality~(\ref{eq_738569.5889}), using (\ref{eq_738570.8644}), can be expressed as
\begin{IEEEeqnarray}{rCl}
\label{eq_738573.4626}\frac{\dd \mathcal{R}(\beta^{-1})}{\dd \beta^{-1}}  & < & \frac{2}{pN} \frac{pN\mathcal{R}(\beta^{-1})-\sum_{j=1}^p  \|  b_j^2 \|^2}{\tau +\beta^{-1}} \\
\label{eq_738575.4651}& < & \frac{2}{pN} \frac{\sum_{j=1}^p  \|  b_j^1 \|^2}{\tau +\beta^{-1}},
  \end{IEEEeqnarray}
where (\ref{eq_738575.4651}) follows using (\ref{eq_738575.4658}). Inequalities (\ref{eq_738573.4614}) and (\ref{eq_738573.4626}) lead to (\ref{eq_738573.4562}).
\paragraph{Part 3} 
Introduce $h(\beta^{-1})  =  \mathcal{R}(\beta^{-1}) - \beta^{-1}$, and observe that $h(\beta^{-1}|_{low}) > 0$ and $h(\beta^{-1}|_{up}) < 0$. By the intermediate value theorem, there is a $\hat{\beta}^{-1} \in \left(\beta^{-1}|_{low}, \beta^{-1}|_{up}  \right)$ such that $h(\hat{\beta}^{-1}) = 0$, i.e., $\hat{\beta}^{-1}  =\mathcal{R}(\hat{\beta}^{-1})$. Thus, $\hat{\beta}^{-1}$ is a fixed point of $\mathcal{R}(\beta^{-1})$.    

\section{Proof of Result~\ref{result_convergence}}\label{appendix_738612.6484}
As $\beta^{-1}|_{it} > \beta^{-1}|_{low}$, it follows from~(\ref{eq_738575.721}) that
\begin{IEEEeqnarray}{rCl}
\tau + \beta^{-1}|_{it} & > &  \frac{2}{pN}\sum_{j=1}^p  \| \mathrm{y}_j  \|^2,
\end{IEEEeqnarray} 
and thus
\begin{IEEEeqnarray}{rCCCCl}
 \frac{\dd \mathcal{R}(\beta^{-1}|_{it})}{\dd \beta^{-1}}  & < & \frac{\sum_{j=1}^p  \|  b_j^1 \|^2}{\sum_{j=1}^p  \| \mathrm{y}_j  \|^2} 
& = & \frac{\sum_{j=1}^p  \|  b_j^1 \|^2}{\sum_{j=1}^p (\|  b_j^1 \|^2 +  \|  b_j^2 \|^2)}.  \IEEEeqnarraynumspace 
  \end{IEEEeqnarray}  
That is, there exists a constant $k$ such that  
\begin{IEEEeqnarray}{rCCCCCl}
\label{eq_738570.5074}0 & < & \frac{\dd \mathcal{R}(\beta^{-1}|_{it})}{\dd \beta^{-1}}  & \leq & k & < 1,\; \forall it \in \{0,1,2,\cdots \}.
  \end{IEEEeqnarray}  
Let $\hat{\beta}^{-1}$ be a fixed point of $\mathcal{R}(\beta^{-1})$. Now, consider
 \begin{IEEEeqnarray}{rCl} 
 \left | \beta^{-1}|_{it} -  \hat{\beta}^{-1} \right | & = &  \left | \mathcal{R}(\beta^{-1}|_{it-1}) - \mathcal{R}(\hat{\beta}^{-1}) \right | \\
  & \leq & k \left | \beta^{-1}|_{it-1} -  \hat{\beta}^{-1} \right | \\
 && \vdots \\
  & \leq & k^{it} \left | \beta^{-1}|_0 - \hat{\beta}^{-1} \right |,
   \end{IEEEeqnarray}  
 that leads to 
 \begin{IEEEeqnarray}{rCCCl} 
\lim_{it \to \infty} \left | \beta^{-1}|_{it} -  \hat{\beta}^{-1} \right | & \leq & \lim_{it \to \infty} k^{it} \left | \beta^{-1}|_0 -  \hat{\beta}^{-1} \right | & = & 0. \IEEEeqnarraynumspace
    \end{IEEEeqnarray}  
The uniqueness of the fixed point can be seen via assuming by contradiction that there exists another fixed point, say $\tilde{\beta}^{-1}$. Now consider
 \begin{IEEEeqnarray}{rCCCCl} 
 \left | \tilde{\beta}^{-1} -  \hat{\beta}^{-1} \right | & = & \left |\mathcal{R}(\tilde{\beta}^{-1}) -  \mathcal{R}(\hat{\beta}^{-1}) \right | & \leq &  k \left | \tilde{\beta}^{-1} -  \hat{\beta}^{-1} \right | \IEEEeqnarraynumspace \\
 &&& < & \left | \tilde{\beta}^{-1} -  \hat{\beta}^{-1} \right |. \IEEEeqnarraynumspace
    \end{IEEEeqnarray}  
This implies that $\tilde{\beta}^{-1}  =  \hat{\beta}^{-1} $. Hence, the result follows. 

\section{Proof of Result~\ref{result_738582.4835}}\label{appendix_738614.4953}
According to the triangle inequality,     
 \begin{IEEEeqnarray}{rCl}  
\nonumber \lefteqn{\left \| (K_{\mathrm{x}\mathrm{a}} + \Delta_x K_{\mathrm{a}\mathrm{a}}^{1/2})\alpha_j^* - ( \mathrm{y}_j + \Delta \mathrm{y}_j ) \right \|} \\
\nonumber  & \leq &   \left \| K_{\mathrm{x}\mathrm{a}} \alpha_j^* - \mathrm{y}_j  \right \| + \left \| \Delta_x K_{\mathrm{a}\mathrm{a}}^{1/2} \alpha_j^* -  \Delta \mathrm{y}_j  \right \| \\
\nonumber & \leq & \left \| K_{\mathrm{x}\mathrm{a}} \alpha_j^* - \mathrm{y}_j  \right \| + \left \| \left[\begin{IEEEeqnarraybox*}[][c]{,c/c,} \Delta_x &  \Delta \mathrm{y}_j\end{IEEEeqnarraybox*} \right] \right \|_F \left \| \left[\begin{IEEEeqnarraybox*}[][c]{,c,} K_{\mathrm{a}\mathrm{a}}^{1/2} \alpha_j^* \\  -1 \end{IEEEeqnarraybox*} \right] \right \| \IEEEeqnarraynumspace \\
& \leq & \left \| K_{\mathrm{x}\mathrm{a}} \alpha_j^* - \mathrm{y}_j  \right \| + \delta_{m} \sqrt{1+ (\alpha_j^*)^T  K_{\mathrm{a}\mathrm{a}} \alpha_j^*},
 \end{IEEEeqnarray}   
 and hence
 \begin{IEEEeqnarray}{rCl}  
\label{eq_738582.4045} \hat{\alpha}_j & = & \left( K_{\mathrm{x}\mathrm{a}}^T K_{\mathrm{x}\mathrm{a}} + \delta_{m} \frac{ \left \| K_{\mathrm{x}\mathrm{a}} \hat{\alpha}_j - \mathrm{y}_j  \right \|}{\sqrt{1+ \hat{\alpha}_j^TK_{\mathrm{a}\mathrm{a}} \hat{\alpha}_j }}K_{\mathrm{a}\mathrm{a}} \right)^{-1} K_{\mathrm{x}\mathrm{a}}^T \mathrm{y}_j. \IEEEeqnarraynumspace
  \end{IEEEeqnarray} 
Algorithm~\ref{algorithm_convergent_basic_learning} estimates $\alpha_j(\hat{\beta}^{-1})$ using (\ref{eq_vector_alpha}). It  can be seen using (\ref{eq_vector_alpha}) that 
  \begin{IEEEeqnarray}{rCl}  
\label{eq_738582.3946} (\tau + \hat{\beta}^{-1}) \frac{\sqrt{1+(\alpha_j(\hat{\beta}^{-1}))^TK_{\mathrm{a}\mathrm{a}}\alpha_j(\hat{\beta}^{-1}) }}{ \|  K_{\mathrm{x}\mathrm{a}} \alpha_j(\hat{\beta}^{-1}) - \mathrm{y}_j   \| } & = & \delta_{m}.
  \end{IEEEeqnarray}  
As a result of (\ref{eq_738582.3946}), it follows from (\ref{eq_vector_alpha}) that
 \begin{IEEEeqnarray}{rCl}  
\label{eq_738582.4003} \lefteqn{ \alpha_j(\hat{\beta}^{-1})  =  \left( K_{\mathrm{x}\mathrm{a}}^T K_{\mathrm{x}\mathrm{a}} \right.} \\
\nonumber && \left. {+}\: \delta_{m} \frac{ \|  K_{\mathrm{x}\mathrm{a}} \alpha_j(\hat{\beta}^{-1}) - \mathrm{y}_j   \|}{\sqrt{1+(\alpha_j(\hat{\beta}^{-1}))^TK_{\mathrm{a}\mathrm{a}}\alpha_j(\hat{\beta}^{-1}) }}   K_{\mathrm{a}\mathrm{a}} \right )^{-1}  K_{\mathrm{x}\mathrm{a}}^T \mathrm{y}_j. \IEEEeqnarraynumspace
  \end{IEEEeqnarray}  
As equalities (\ref{eq_738582.4003}) and (\ref{eq_738582.4045}) are identical, $\alpha_j(\hat{\beta}^{-1})$ (which is the solution of (\ref{eq_738582.4003})) must be equal to $\hat{\alpha}_j$ (which is the solution of (\ref{eq_738582.4045})), i.e.,
 \begin{IEEEeqnarray}{rCl}  
 \alpha_j(\hat{\beta}^{-1}) & = & \hat{\alpha}_j.
  \end{IEEEeqnarray}  
Hence, Algorithm~\ref{algorithm_convergent_basic_learning} solves the min-max problem~(\ref{eq_738582.4785}).  
}

\bibliographystyle{IEEEtran}

\bibliography{IEEEabrv,bibliography}

\vfill

\end{document}